%% file: acl_latex.tex
\documentclass[11pt]{article}

\usepackage[preprint]{acl}
\usepackage{tcolorbox}
\tcbuselibrary{breakable,listings,skins}
\usepackage{subcaption}

\usepackage{times}
\usepackage{latexsym}

\usepackage[T1]{fontenc}

\usepackage[utf8]{inputenc}

\usepackage{microtype}

\usepackage{inconsolata}

\usepackage{graphicx}
\usepackage{amsmath}
\usepackage{algorithm}
\usepackage{algpseudocode}
\usepackage{array}
\usepackage{enumitem}
\usepackage{booktabs}
\usepackage{multirow}
\usepackage{xcolor}
\usepackage{colortbl}
\usepackage{ragged2e}
\usepackage{tikz}
\usepackage{pgfplots}
\pgfplotsset{compat=1.18}
\usepgfplotslibrary{groupplots}
\usetikzlibrary{arrows.meta, positioning, calc, shadows.blur, fit, backgrounds,
                decorations.pathmorphing, shapes.geometric}
\definecolor{blue1}{HTML}{001A6E}
\definecolor{sceneFieldsColor}{HTML}{B66A4A}
\usetikzlibrary{arrows.meta,calc,fit,positioning,shapes.symbols}

\definecolor{headerblue}{RGB}{235,242,250}
\definecolor{bestgold}{RGB}{255,250,230}
\definecolor{bucolor}{HTML}{2166AC}
\definecolor{tdcolor}{HTML}{D6604D}
\definecolor{promptcolor}{HTML}{2E7D8C}
\definecolor{prompthl}{HTML}{D4860A}

\newtcblisting{promptbox}[1][]{
  enhanced,
  breakable,
  listing only,
  colback=white,
  colframe=promptcolor,
  colbacktitle=promptcolor,
  coltitle=white,
  fonttitle=\bfseries\footnotesize,
  halign title=flush center,
  left=4pt,
  right=4pt,
  top=4pt,
  bottom=4pt,
  boxsep=1.5pt,
  arc=2pt,
  before skip=6pt,
  after skip=6pt,
  listing options={
    basicstyle=\ttfamily\scriptsize,
    columns=fullflexible,
    keepspaces=true,
    escapeinside={(*@}{@*)},
    showstringspaces=false,
    breaklines=true,
    breakindent=0pt,
    breakatwhitespace=true,
    postbreak={},
    tabsize=2,
    upquote=true,
    xleftmargin=0pt,
    frame=none,
    aboveskip=0pt,
    belowskip=0pt,
  },
  #1
}

\newtcolorbox{tracebox}[2][]{
  enhanced,
  breakable,
  colback=white,
  colframe=sceneFieldsColor,
  colbacktitle=sceneFieldsColor,
  coltitle=white,
  fonttitle=\bfseries\footnotesize,
  halign title=flush center,
  left=4pt,
  right=4pt,
  top=4pt,
  bottom=4pt,
  boxsep=2pt,
  arc=2pt,
  before skip=5pt,
  after skip=5pt,
  title={#2},
  #1
}
\newtcolorbox{tracefloatbox}[2][]{
  enhanced,
  colback=white,
  colframe=sceneFieldsColor,
  colbacktitle=sceneFieldsColor,
  coltitle=white,
  fonttitle=\bfseries\footnotesize,
  fontupper=\small,
  halign title=flush center,
  left=4pt,
  right=4pt,
  top=4pt,
  bottom=4pt,
  boxsep=2pt,
  arc=2pt,
  title={#2},
  #1
}
\newcommand{\promptcaption}[2]{%
  \nopagebreak[4]%
  \noindent\begin{minipage}{\linewidth}
    \captionsetup{hypcap=false,font=small,skip=2pt}%
    \captionof{table}{#1}%
    \label{#2}%
  \end{minipage}%
  \par\medskip
}
\title{Storyline Trees: Hierarchical Representations for Long-Form Narratives}

\author{
  Litu Ou\\
  School of Informatics\\
  University of Edinburgh\\
  \texttt{litu.ou@ed.ac.uk}
  \And
  Mirella Lapata\\
  School of Informatics\\
  University of Edinburgh\\
  \texttt{mlap@inf.ed.ac.uk}
}

\begin{document}
\maketitle
\begin{abstract}

Long-form narratives are challenging for long-context models because their  structure is implicit: events, characters, and plotlines interact across hundreds of pages without the explicit  cues that guide navigation in structured documents. We address this by constructing \emph{storyline trees}, hierarchical representations that organize narratives from global themes and major plotlines to fine-grained events. We first segment chapters into contiguous narrative segments, or \emph{scenes}, and use them as the basic units for tree construction. We then infer storyline trees through complementary top-down and bottom-up procedures that derive, refine, cluster, and summarize storylines at multiple levels of abstraction. We showcase the utility of this representation for question answering: storyline trees enable \emph{adaptive retrieval}, allowing models to iteratively inspect high-level narrative structure and retrieve scene-level evidence on demand. Experiments on three long-context narrative QA benchmarks show that adaptive retrieval outperforms strong baselines, including post-trained long-context models and agentic chunk-based methods. Ablations confirm that scenes are more effective basic units than chapters or generic segmentation, and that gains persist under matched retrieval budgets\footnote{Code available in \url{https://github.com/Leonard907/StoryTree}}.

\end{abstract}

\section{Introduction}
\label{sec:introduction}

Long-form narratives such as novels have proven challenging for
current long-context language models
\citep{kryscinski2022booksum,wang2025novelqa,du2025context}. Even when
the entire text fits within a model's context window, performance
degrades sharply on tasks that require identifying and integrating
information dispersed across hundreds of pages
\citep{tian2025distance,veseli2025positional,xu2025divideconquer}. Unlike
technical writing, which makes its structure explicit through devices
such as headings, narrative writing dramatizes events without
commentary, relying on the reader to infer the connections. Characters'
motives  and causal links between events may be distributed across hundreds of pages without any direct linguistic signal. Chapter boundaries are often dictated by length, pacing or stylistic choice rather than thematic coherence, making them insufficient for tracing deeper narrative threads.

Existing methods are largely designed for documents whose organization
is explicitly marked. One line of work builds hierarchies from section
headings, document layout, or discourse relations
\citep{jin2025hierarchicalrefinement,zhou2025contextedus,zhang2026docdancer}.
These signals are useful for navigating structured documents, but they
do not recover the implicit narrative threads that organize events
across a story. A second line of work is more directly applicable to
narratives: it
decomposes text into generic segments and processes them by
incrementally updating a compact memory or by exchanging messages among
chunk-level agents
\citep{zhang2024coa,yu2025memagent,yu2025treeagents}. Such methods can
aggregate evidence across a document, but their intermediate states are
typically tied to a \emph{specific} question or inference trajectory,
rather than forming a reusable representation of the narrative. 



\begin{figure}[t]
    \centering
    \setlength{\abovecaptionskip}{2pt}
    \setlength{\belowcaptionskip}{-3pt}
    \includegraphics[width=\columnwidth,trim=160bp 270bp 160bp 270bp,clip]{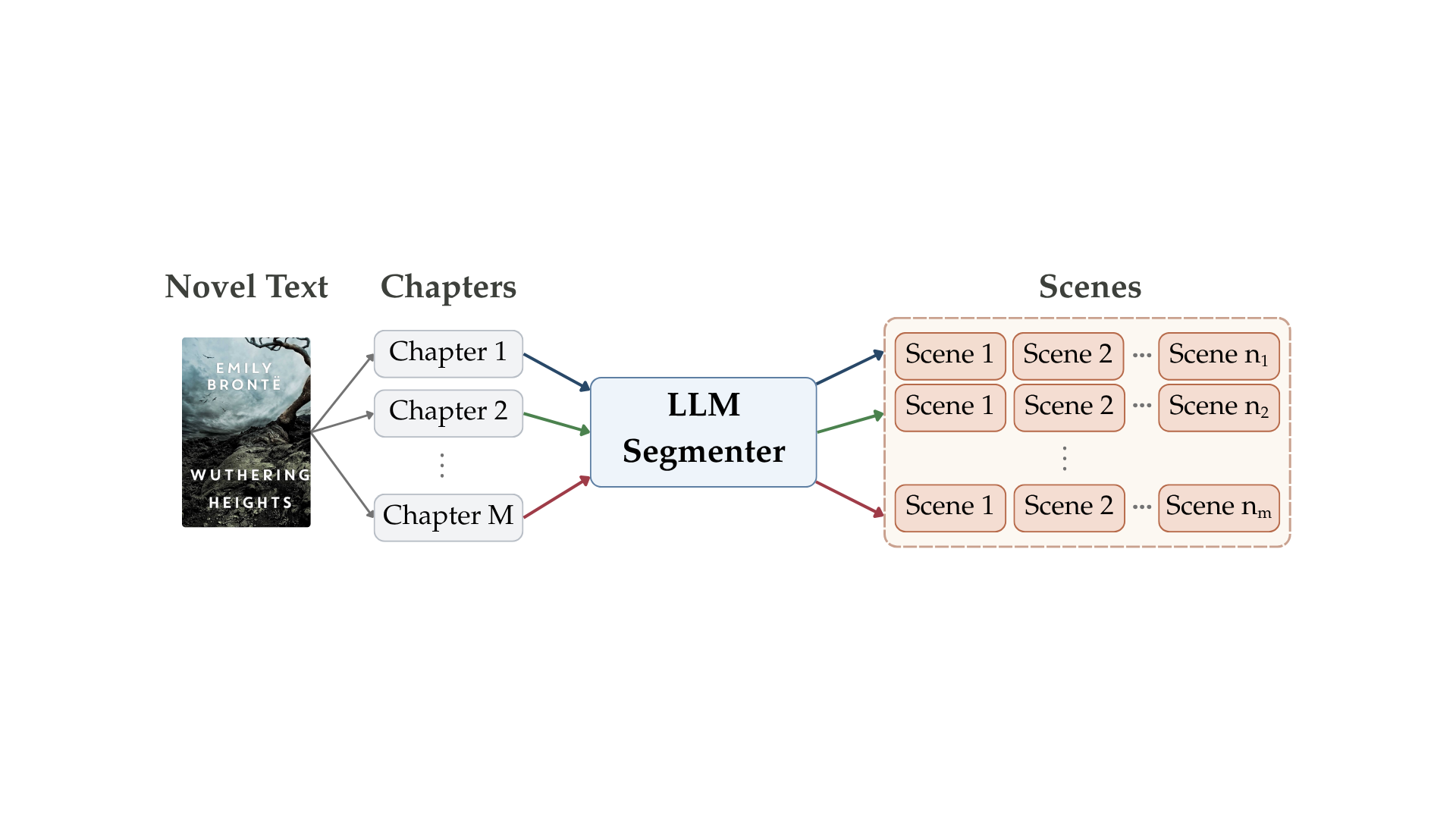}
    \vspace{-0.8em}
    \caption{Scene segmentation: a book is first divided into chapters
    using metadata; each chapter is then passed to an LLM, which
    segments it into scenes that serve as the basic units for
    storyline-tree construction.}
    \label{fig:scene-breakdown}
    \vspace{-0.8em}
\end{figure}

We posit that long-form narrative QA benefits from modeling implicit storyline structure rather than treating narratives as flat collections of retrievable passages; to this end, we introduce \emph{storyline trees}, hierarchical representations that organize long-form narratives from global themes and major plotlines down to fine-grained events. We first segment each narrative into \emph{scenes} (see Figure~\ref{fig:scene-breakdown}), which serve as the fundamental units of representation and provide a consistent level of granularity across different narratives.
We then infer storyline trees following two
complementary views of narrative organization. \emph{Top-down}
construction first infers abstract storylines from the full set of
scenes and recursively refines them into more specific sub-storylines
(see Figure~\ref{fig:topdown}). It is aligned with a plot-
or theme-first view, in which individual events acquire meaning through
their role in larger narrative arcs
\citep{chatman1978story,ryan2005narrative}. \emph{Bottom-up}
construction recursively clusters and summarizes scenes into
increasingly abstract storyline nodes until no further clustering is
possible (see Figure~\ref{fig:bottomup}). It is aligned with a
scene- or event-first view, in which broader storylines emerge from
local relations among scenes, such as shared characters, causal links,
recurring motifs, or temporal progression
\citep{bal1997narratology,zwaan1998situationmodels,cohn2013visual}. Although
the two approaches induce trees in opposite directions, they yield the
same abstraction order: higher-level nodes capture broad storyline
structure, while lower-level nodes ground these abstractions in
individual scenes.

We showcase the utility of storyline trees on narrative question
answering by developing an \emph{adaptive retrieval} procedure. Given a
question, a QA model first retrieves an initial set of scenes, then
iteratively uses the storyline tree to decide which parts of the
narrative to inspect next. At each step, it may request additional
evidence from a storyline node or produce an answer from the evidence
accumulated so far. Our representation allows retrieval decisions to be
made at multiple levels of abstraction, enabling the model to
progressively narrow its focus from broad storyline structure to
localized scene evidence rather than treating the narrative as a flat
collection of passages.

Experimental results across three long-context narrative QA datasets demonstrate that adaptive retrieval over storyline trees consistently achieves the best performance. Further analysis shows that both scene-based construction and adaptive retrieval contribute to these gains, and that storyline trees are especially effective for questions that require locating and interpreting narrative events within their broader story context. Our contributions are: 
\begin{itemize}[itemsep=0pt,labelindent=0pt,leftmargin=*]
    \item We propose storyline trees, an interpretable hierarchical
    representation for long-form narratives that connects global
    themes and plotlines with fine-grained scene-level evidence.
    \item We introduce top-down and bottom-up tree construction
    methods, together with an adaptive retrieval procedure that allows
    models to progressively explore the tree when answering questions.
    \item We demonstrate improvements on three long-context narrative
    QA benchmarks and provide ablations showing the effectiveness of
    scenes, tree structure, and adaptive traversal.
\end{itemize}

\section{Related Work}

\paragraph{Narrative Units and Scene Structure}
Prior work in narrative comprehension provides a theoretical basis for
defining units around changes in the described situation rather than
around surface form or fixed length. Situation-model and
event-perception accounts argue that readers track dimensions such as
time, space, characters, and goals, updating their mental
representation of the narrative when these dimensions shift
\citep{zwaan1995eventindexing,zwaan1998situationmodels,zacks2007eventperception}.
Computational work operationalizes this intuition by defining scenes as
coherent stretches of narrative with consistent time, location,
character configuration, and action
\citep{gaizauskas2015sceneml,zehe2021detectingscenes}. In screenplay
analysis, scenes have also served as textual anchors for mapping
higher-level plot structure \citep{papalampidi2019movieplot}. Taken
together, these studies support our decision to segment novels into
scenes.

\paragraph{Long-Context Narrative QA and Structured Retrieval}
A large body of work shows that increased context windows do not by
themselves yield reliable long-context reasoning: model performance
remains sensitive to where evidence appears in the context, how long
the input is, and how much the question and evidence lexically overlap
\citep{liu2023lostinmiddle,hsieh2024ruler,modarressi2025nolima,du2025context}.
These limitations are particularly pronounced for long-form literary
narratives, whose interpretation requires tracking events, characters,
and their motivations across extended discourse
\citep{kryscinski2022booksum,xu2025detectiveqa,wang2025novelqa,bonomo2025literaryqa}.
These studies collectively suggest that narrative understanding
requires access to fine-grained events \emph{and} higher-level
reasoning over plotlines and themes.

Structured approaches to long-context retrieval fall into two broad
categories. Hierarchical methods organize documents into coarse-to-fine
representations for multi-level retrieval or refinement
\citep{zhou2025contextedus,jin2025hierarchicalrefinement,tao2025treerag},
but are designed around explicit document structure and do not transfer
naturally to narratives. Agentic approaches aggregate information
through memory updates or multi-agent communication over segmented
inputs \citep{zhang2024coa,yu2025memagent,yu2025treeagents}, but their
intermediate states are tied to a specific question rather than forming
a reusable representation. Our storyline tree is closest to
hierarchical retrieval, but is specialized for narrative structure:
leaves correspond to scenes and internal nodes encode plotlines and
themes, enabling narrative-aware retrieval that generic representations
(e.g., chunks or summaries) do not support.

\section{Constructing Storyline Trees}
\label{sec:methodology}
\label{sec:tree-construction}



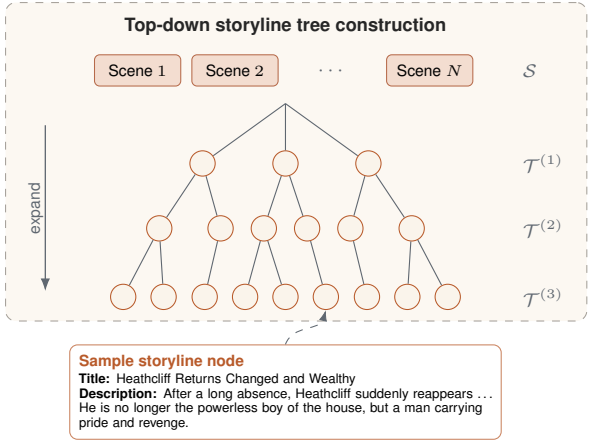
\begin{figure}[t]
    \centering
    \setlength{\abovecaptionskip}{2pt}
    \setlength{\belowcaptionskip}{-3pt}
    \input{topdown2.tex}
    \vspace{-0.8em}
    \caption{Top-down tree induction infers abstract storylines
    from the full set of scenes and recursively refines them into more
    specific sub-storylines.}
    \label{fig:topdown}
    \vspace{-0.8em}
\end{figure}

\subsection{Scenes as Basic Narrative Units}

A key first step is to define the basic unit over which storyline
trees are constructed. Chapters may seem like a natural candidate, as
they are the author's intended segmentation. However, they are often
poorly aligned with narrative structure. Chapter boundaries may be
shaped by pacing or stylistic considerations, and their granularity
varies substantially across works: a chapter may contain several
storylines in one novel while covering a much narrower narrative unit
in another. Consequently, chapters provide an inconsistent basis for
representation, being too coarse for some narratives and too
fine-grained for others.

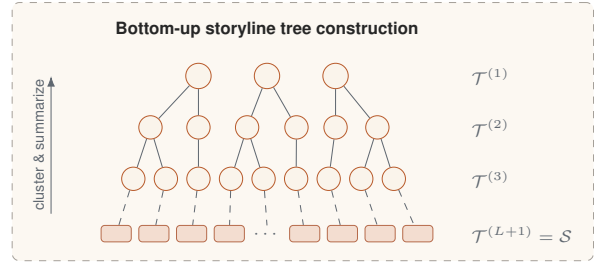
\begin{figure}[t]
    \centering
    \setlength{\abovecaptionskip}{2pt}
    \setlength{\belowcaptionskip}{-3pt}
    \input{bottomup2.tex}
    \vspace{-0.8em}
    \caption{Bottom-up tree induction starts from scene
    representations, recursively clusters related scenes
    and summarizes each cluster into a higher-level storyline node until
    no further clustering is possible.}
    \label{fig:bottomup}
    \vspace{-0.8em}
\end{figure}

An alternative is to partition the text into fixed-length segments.
However, even when these segments preserve sentence or paragraph
boundaries, they can disrupt narrative continuity by separating closely
related events. Semantic chunking seeks to address this limitation by
segmenting text according to semantic coherence rather than length
alone \citep{glavas2016semanticsegmentation}. However, recent work
suggests that its benefits are inconsistent and often insufficient to
justify the additional computational cost
\citep{qu2024semanticchunking}. More importantly, semantic chunking
does not explicitly model the narrative structures that are central to
novels. This motivates a segmentation strategy tailored specifically to
long-form narrative text.

We therefore use \emph{scenes} as the basic unit for storyline trees.
In narrative theory, a scene is a contiguous segment characterized by
relative consistency in time, location, and narrative focus; a boundary
is introduced when one of these dimensions changes
\citep{santana2023survey,papalampidi2019movieplot}. Figure~\ref{fig:scene-breakdown}
illustrates our scene extraction pipeline. We parse the source text
into chapters and then apply a standardized scene extraction prompt to
segment each chapter into scenes. Using the same prompt across novels
encourages more consistent segmentation across books and reduces
variation introduced by differences in chapter structure and writing
style. In addition to identifying scene boundaries, the LLM generates
structured descriptions for each scene.

For each scene, the model produces a title, a synopsis, and a
significance statement; an example output is provided in
Table~\ref{tab:scene-fields}. The \emph{Title} provides a
concise label for the scene, the \emph{Synopsis} summarizes its
content, and the \emph{Significance} explains its narrative function or
implicit meaning. To make scene boundaries recoverable in the original
text, the prompt also returns the final sentence of each scene as a
boundary marker. Together, these fields provide compact scene-level
summaries with meaningful annotations for subsequent tree construction.

\subsection{Top-Down Tree Induction}
\label{sec:top-down}

Top-down tree induction begins by inferring high-level narrative
themes from the full collection of scenes and then recursively
decomposing them into increasingly specific sub-storylines (see
Figure~\ref{fig:topdown}). Our algorithm operates over structured
scene representations rather than the original narrative text. Each
storyline node at level~$l$ is represented as
\begin{equation}
t_k^{(l)} = \big(\tau_k^{(l)},\, \delta_k^{(l)}\big),
\label{eq:storyline-node}
\end{equation}
where $\tau_k^{(l)}$ is a short title and $\delta_k^{(l)}$ is a
textual description of the storyline (see Figure~\ref{fig:topdown}).

Let $\mathcal{S}=\{s_i\}_{i=1}^N$ denote the ordered set of extracted
scenes, where each scene $s_i$ consists of its title, synopsis, and
significance statement (cf.~Table~\ref{tab:scene-fields}). We write
$\mathcal{X}=[s_1;\ldots;s_N]$ for their concatenation in narrative
order. The top-down procedure first generates the most abstract
storyline nodes from $\mathcal{X}$, then recursively generates
each subsequent level by conditioning on  $\mathcal{X}$ and the
complete set of storyline nodes from the previous level:
\begin{equation}
\mathcal{Y}^{(l)} =
\begin{cases}
f_{\mathrm{TD}}(\mathcal{X}), & l = 1, \\[2pt]
f_{\mathrm{TD}}(\mathcal{X}, \mathcal{T}^{(l-1)}), & 2 \leq l \leq L,
\end{cases}
\label{eq:td-generation}
\end{equation}
where $f_{\mathrm{TD}}$ denotes the LLM-based top-down generation
function, and $\mathcal{Y}^{(l)}$ denotes the structured records
generated at level~$l$. For $l=1$, each generated record contains a
title and a description. For $l>1$, each generated record additionally
contains parent references indicating which nodes in
$\mathcal{T}^{(l-1)}$ it elaborates.

For $l=1$, the generated records directly define the first-level
storyline nodes:
\begin{equation}
\mathcal{T}^{(1)} = \{(\tau_k^{(1)}, \delta_k^{(1)}) :
y_k^{(1)} \in \mathcal{Y}^{(1)}\}.
\end{equation}
For each subsequent level, generation is performed jointly over the complete set of the parent storylines from the previous level. We write each generated record as
\begin{multline}
y_k^{(l)} =
\big(\tau_k^{(l)},\, \delta_k^{(l)},\, \mathcal{P}_k^{(l)}\big),\\
y_k^{(l)} \in \mathcal{Y}^{(l)}, \quad 2 \leq l \leq L,
\label{eq:td-record}
\end{multline}
where
\(\mathcal{P}_k^{(l)} \subseteq \mathcal{T}^{(l-1)}\)
is the set of parent storylines from the previous level that the
generated record elaborates. This allows a sub-storyline to refine a
single parent storyline or to connect multiple related parent
storylines.

The corresponding storyline nodes are 
\begin{equation}
\mathcal{T}^{(l)} =
\{t_k^{(l)} = (\tau_k^{(l)}, \delta_k^{(l)}) :
y_k^{(l)} \in \mathcal{Y}^{(l)}\}.
\label{eq:td-nodes}
\end{equation}
The parent references induce the edge set between adjacent levels:
\begin{equation}
\begin{aligned}
\mathcal{E}^{(l)}
&=
\big\{(p, t_k^{(l)}) :
y_k^{(l)} \in \mathcal{Y}^{(l)},\,
p \in \mathcal{P}_k^{(l)}\big\},\\[-1pt]
&\hfill 2 \leq l \leq L.
\end{aligned}
\label{eq:td-edges}
\end{equation}
Conditioning on~$\mathcal{X}$ keeps each generation step grounded in
the full sequence of scene-level evidence, while conditioning on the
complete previous storyline level~$\mathcal{T}^{(l-1)}$ provides a
structural guide that constrains the model to elaborate existing
storylines into more specific sub-storylines rather than introduce
unrelated narrative themes.

\subsection{Bottom-Up Tree Induction}
\label{sec:bottom-up}
Inspired by \citet{sarthi2024raptor}, the second approach induces the
storyline tree in a bottom-up manner. It begins at the scene level,
treating each scene as an atomic unit. As illustrated in
Figure~\ref{fig:bottomup}, semantically related scenes are first
clustered, and each cluster is summarized into a storyline node
following the title--description schema in
Equation~\eqref{eq:storyline-node}. The same cluster-and-summarize
operation is then applied recursively to the generated storyline
nodes, producing increasingly abstract tree levels.


We index the resulting tree so that $\mathcal{T}^{(1)}$ denotes the
most abstract level and $\mathcal{T}^{(L)}$ denotes the lowest
storyline level, immediately above the scene level. Let
$\mathcal{S}=\{s_i\}_{i=1}^{N}$ denote the ordered set of extracted
scenes. To make the recursion uniform, we treat the scenes as a
virtual level just below the tree by setting
$\mathcal{T}^{(L+1)} = \mathcal{S}$.

The bottom-up procedure recursively maps a set of lower-level units
to a set of higher-level storyline nodes. For each level~$l$, where
$1 \leq l \leq L$, the cluster-and-summarize operation first
partitions the units in $\mathcal{T}^{(l+1)}$ into clusters:
\begin{equation}
\hspace{-2pt}\mathcal{P}^{(l)}
=
\{P_1^{(l)}, \ldots, P_{K_l}^{(l)}\}
=
\mathrm{Cluster}\big(\mathcal{T}^{(l+1)}\big),
\end{equation}
where each $P_k^{(l)} \subseteq \mathcal{T}^{(l+1)}$ contains
semantically related lower-level units. Each cluster is then
summarized into a higher-level storyline node:
\begin{equation}
t_k^{(l)}
=
\mathrm{Summarize}\big(P_k^{(l)}\big),
\qquad
1 \leq k \leq K_l.
\end{equation}
The set of storyline nodes at level~$l$ is therefore
\begin{equation}
\mathcal{T}^{(l)}
=
\{t_k^{(l)}\}_{k=1}^{K_l}.
\label{eq:bottom-up-induction}
\end{equation}
This formulation makes the tree structure explicit: each higher-level
storyline node is the parent of the lower-level units in the cluster
from which it was summarized. The edge set at level~$l$
is
\begin{equation}
\mathcal{E}^{(l)} =
\big\{(t_k^{(l)},\, u) :
1 \leq k \leq K_l,\, u \in P_k^{(l)}\big\}.
\label{eq:bottom-up-edges}
\end{equation}
The resulting bottom-up storyline tree is
\begin{equation}
\begin{aligned}
\mathcal{G}_{\mathrm{BU}} &= (\mathcal{V},\, \mathcal{E}), \\
\mathcal{V} &= \textstyle\bigcup_{l=1}^{L}\mathcal{T}^{(l)} \cup \mathcal{S}, \\
\mathcal{E} &= \textstyle\bigcup_{l=1}^{L}\mathcal{E}^{(l)}.
\end{aligned}
\label{eq:bottom-up-graph}
\end{equation}
The recursion terminates when the current level can no longer be
meaningfully clustered. The remaining nodes form $\mathcal{T}^{(1)}$,
corresponding to the most abstract thematic level of the tree.
Although the top-down and bottom-up approaches proceed in opposite
directions, both use the same abstraction order:
$\mathcal{T}^{(1)}$ captures broad narrative themes, while deeper
levels represent  more specific content.

\section{Storyline-Guided Question Answering}
\label{sec:qa-tasks}

We next explain how storyline trees support question answering.
Existing methods organize long documents hierarchically and retrieve
relevant units before answer generation
\citep{tao2025treerag,jin2025hierarchicalrefinement,zhou2025contextedus}.
Although this improves over flat retrieval by exposing different levels
of granularity, the retrieval or compression policy is typically fixed
in advance. The resulting evidence set may therefore omit important
information needed for questions whose support is distributed across
the narrative.

To address this limitation, we propose \emph{adaptive retrieval}, summarized in Algorithm~\ref{alg:adaptive-retrieval}. The procedure operates on storyline trees $\mathcal{G} \in \{\mathcal{G}_{\mathrm{TopDown}}, \mathcal{G}_{\mathrm{BottomUp}}\}$ produced by either induction method. Given a question~$q$, we first retrieve an
initial evidence set $\mathcal{R}$ from the scene pool $\mathcal{S}$
using $q$ as the query. For retrieval, each scene in~$\mathcal{S}$ is represented by its full
original source text rather than by the LLM-generated structured summary fields (Title, Synopsis and Significance). 
 Dense embeddings for the scene pool are
computed from the original scene text. 
At each subsequent step, we provide the LLM
with the question, a serialized representation of $\mathcal{G}$, and
the currently retrieved scenes; the model then returns a pair
$(a_r,p_r)$ in which the action $a_r$ either commits to an answer or
requests more evidence. The payload $p_r$ is the predicted answer when
$a_r=\textsc{Answer}$, and a node identifier otherwise. In the latter
case, $\operatorname{Resolve}$ maps the identifier to the corresponding
node $v_r$ of $\mathcal{G}$, and the system uses the node's title
$\tau(v_r)$ as a query to retrieve the top-$k$ most relevant scenes
from $\mathcal{S}$ via the same retriever used for the initial query.
The newly retrieved scenes are merged into $\mathcal{R}$ by set union,
so already-retrieved scenes are not duplicated, and the loop repeats
until the model emits an answer or the retrieval budget $B$ is
exhausted. If the budget is exhausted before an answer is produced, we
issue a final forced-answer prompt using all evidence collected so far.

This adaptive procedure allows retrieval to be guided by the model's
evolving information needs, rather than being fixed before generation.
The tree's role is to give the LLM an interpretable map of the
narrative from which to select where to look next; the retrieval step
itself is a similarity match against $\mathcal{S}$ rather than an edge
traversal of $\mathcal{G}$. 

\begin{algorithm}[t]
  \caption{Adaptive Storyline-Guided Retrieval}
  \label{alg:adaptive-retrieval}
  \begin{algorithmic}[1]
  \Require Question $q$, storyline tree $\mathcal{G}$, scene pool $\mathcal{S}$, top-$k$, budget $B$
  \State $\mathcal{R} \leftarrow \operatorname{Retrieve}(q, \mathcal{S}, k)$
  \For{$r = 1$ to $B$}
      \State $(a_r, p_r) \leftarrow \operatorname{Decide}_{\mathrm{LLM}}(q, \mathcal{G}, \mathcal{R})$
      \If{$a_r = \textsc{Answer}$}
          \State \Return $p_r$
      \EndIf
      \State $v_r \leftarrow \operatorname{Resolve}(p_r, \mathcal{G})$
      \State $\Delta_r \leftarrow \operatorname{Retrieve}(\tau(v_r), \mathcal{S}, k)$
      \State $\mathcal{R} \leftarrow \mathcal{R} \cup \Delta_r$
  \EndFor
  \State \Return $\operatorname{Answer}_{\mathrm{LLM}}(q, \mathcal{R})$
  \end{algorithmic}
  \end{algorithm}

\section{Experimental Setting}
\label{sec:experiments}

\subsection{Datasets}
\label{sec:datasets}
We evaluate on three book-level narrative QA benchmarks. DetectiveQA tests mystery comprehension over clues, motives, timelines, and character relations~\citep{xu2025detectiveqa}. NovelQA evaluates QA over extremely long inputs and diverse question types~\citep{wang2025novelqa}. LiteraryQA focuses on open-ended literary comprehension  using cleaned QA pairs from NarrativeQA~\citep{bonomo2025literaryqa,kocisky2018narrativeqa}. Details on dataset statistics, preprocessing, and question formats are provided in Appendix~\ref{app:exp-details}.

\subsection{Implementation Details}
\label{sec:implementation-details}

We use models from the Qwen3 family~\citep{yang2025qwen3} for all LLM-based generation components. 
Specifically, we evaluate Qwen3-30B-A3B-Instruct-2507 and Qwen3-235B-A22B-Instruct-2507, using FP8-quantized checkpoints. For dense retrieval, we use Qwen3-Embedding-8B~\citep{zhang2025qwen3embedding}.
%
We fix the depth of  \emph{top-down} to four across all datasets and backbone models. For \emph{bottom-up trees}, unless otherwise specified, we follow the recursive clustering procedure of RAPTOR~\citep{sarthi2024raptor} but modify the summarization step to produce storyline nodes using the title-description schema defined in Section~\ref{sec:tree-construction}. 
We discuss other bottom-up variants in Appendix~\ref{app:bottom-up-variants}. 
%
For adaptive retrieval, we set the parameter~$k$ (see Algorithm~\ref{alg:adaptive-retrieval}) to 20 for the main scene-based experiments and allow at most 10~retrieval rounds. If the round budget is exhausted before the model emits an answer, a forced-answer prompt is issued over the evidence retrieved so far.
More implementation details are provided in Appendix~\ref{app:exp-details}. Relevant prompt templates are provided in Appendix~\ref{app:prompt-template}.

\subsection{Comparison Methods}
\label{sec:compared-methods}


\paragraph{Long-Context Baselines}
\emph{Zero-shot} truncates the input to the maximum context window and
generates an answer directly using the Qwen3-30B-A3B-Instruct-2507 and
Qwen3-235B-A22B-Instruct-2507 FP8-quantized checkpoints.
\textsc{QwenLong-L1.5}~\citep{shen2025qwenlongl15} is a long-context
post-trained variant of Qwen3-30B-A3B-Thinking-2507~\citep{yang2025qwen3}, which belongs to the same Qwen3 family but is a reasoning model rather than an instruct model. We evaluate it with its released inference setup.

\paragraph{Memory-Based Agents}
\emph{MemAgent}~\citep{yu2025memagent} processes the input sequentially, maintaining a compact memory updated after each segment. \textsc{Tree of Agents} (\emph{TOA})~\citep{yu2025treeagents} partitions the input into chunks and uses multiple chunk-level agents to exchange information along tree-structured communication paths.

\paragraph{Retrieval Baselines}
\emph{Static~RAG} retrieves scene-level units using the question as the query and generates an answer from the retrieved scenes. \emph{Collapsed~RAG} pools scenes and internal storyline nodes into a shared retrieval index, enabling retrieval across abstraction levels in a single pass. \emph{Iterative~RAG} issues multiple rounds of retrieval \emph{without} access to the storyline tree, formulating its own queries at each step rather than selecting from explicit storyline nodes.

\paragraph{Basic-Unit Variants}
To isolate the effect of the segmentation strategy, we compare scenes against three alternatives: chapters, TextTiling segments~\citep{hearst1997texttiling}, and breakpoint-based semantic chunks~\citep{qu2024semanticchunking}. All variants follow the same top-down construction pipeline (Figure~\ref{fig:topdown}), with the per-round retrieval budget adjusted for each unit type to approximately match the source-token volume of the 20-scene setting; details are provided in Appendix~\ref{app:segmentation-variants}.

\subsection{Evaluation Protocol}
\label{sec:evaluation}

For multiple-choice questions, no answer options are provided to the model at inference time; we use the full text of the correct option as the evaluation reference. This prevents models from exploiting option-specific cues and unifies evaluation across datasets~\citep{chandak2025answermatching}. Since exact-match and token-level F1 can undercount correct answers expressed with different surface forms~\citep{ho2025extractiveqajudge,badshah2025referenceguidedverdict},
we assess semantic consistency using an LLM-as-a-judge
framework~\citep{zheng2023llmjudge} with
\texttt{gemini-3.1-pro-preview}\footnote{\url{https://deepmind.google/models/gemini/pro/}}
as the evaluator. Models are prompted to produce answers in a
\texttt{\textbackslash boxed\{\}} format~\citep{guo2025deepseekr1} to
facilitate structured answer extraction, and for datasets with
multiple valid references (e.g.,~LiteraryQA), a prediction is judged
correct if it matches any accepted answer.

\section{Results and  Analysis}
\label{sec:results}

\begin{table}[t]
  \centering
  \small
  \setlength{\abovecaptionskip}{2pt}
  \setlength{\belowcaptionskip}{-3pt}
  \setlength{\tabcolsep}{5pt}
  \begin{tabular}{@{}l@{~}ccc@{}}
    \toprule
     \rowcolor{headerblue}
    \textbf{Method} & \textbf{DetectiveQA} & \textbf{NovelQA} & \textbf{LiteraryQA} \\
    \midrule
    \multicolumn{4}{c}{\textit{Qwen3-30B-A3B-Instruct-2507}} \\
    \midrule
    Zero-shot       & 30.87 & 48.78 & 56.48 \\
    MemAgent        & 27.35 & 40.06 & 52.65 \\
    TOA             & 32.72 & 45.71 & 58.25 \\
    QwenLong-L1.5   & 32.21 & 45.34 & 57.76 \\
    \midrule
    Bottom-up (ours) & \underline{34.06} & \underline{50.99} & \underline{58.64} \\
    Top-down (ours)  & \textbf{35.91}   & \textbf{51.45}   & \textbf{59.51}   \\
    \midrule
    \multicolumn{4}{c}{\textit{Qwen3-235B-A22B-Instruct-2507}} \\
    \midrule
    Zero-shot       & 33.72 & 51.81 & 60.24 \\
    MemAgent        & 30.03 & 56.32 & 61.21 \\
    TOA             & 38.76 & 58.06 & \underline{68.14} \\
    QwenLong-L1.5   & --    & --    & --    \\
    \midrule
       \rowcolor{bestgold}
    Bottom-up (ours)) & \underline{39.60} & \underline{59.39} & 67.72            \\   \rowcolor{bestgold}
    Top-down (ours)  & \textbf{40.27}   & \textbf{60.27}   & \textbf{69.19}   \\
    \bottomrule
  \end{tabular}
  \caption{Accuracy on DetectiveQA, NovelQA, and LiteraryQA with (FP8-quantized) Qwen3 backbones. For \textsc{QwenLong-L1.5}, we report results for the released 30B-A3B checkpoint (no 235B-A22B checkpoint is available). Best results in each dataset-backbone setting are in \textbf{bold}; second best results are \underline{underlined}.}
  \label{tab:qa-results}
  \vspace{-0.8em}
\end{table}


\paragraph{Do storyline trees outperform long-context and agentic baselines?}
Yes, consistently. Across all six dataset-backbone combinations in
Table~\ref{tab:qa-results}, top-down adaptive retrieval achieves the
best accuracy, outperforming zero-shot, MemAgent, TOA, and, where available, 
QwenLong-L1.5. The pattern holds for both the 30B and 235B backbones,
indicating that storyline trees provide a complementary structural
benefit rather than simply compensating for limited model capacity.
Significance tests (Appendix~\ref{app:significance}) show the clearest
support on NovelQA and consistently positive effects elsewhere.

\paragraph{Does construction direction matter?}
Top-down construction is consistently stronger than bottom-up across
all datasets and both model scales, while bottom-up trees are
themselves competitive with strong baselines. This suggests that
hierarchical organization is generally beneficial for long-form
narrative QA, but that imposing a global narrative structure first and
refining it downward yields a better retrieval interface than
aggregating locally.

\paragraph{Are gains uniform across datasets?}
The advantage is most pronounced on DetectiveQA and NovelQA, where answering questions frequently requires locating and connecting evidence distributed across the narrative---precisely the setting where storyline-guided traversal is most useful. On LiteraryQA, relative gains are smaller, consistent with its greater emphasis on broad plot
comprehension and more flexible answer realizations.

\paragraph{Are scenes the right basic unit?}
Yes. Table~\ref{tab:ablation-units} shows that
scene-based construction outperforms all alternatives across all 
three datasets under an approximately matched token-level retrieval budget. The advantage
over chapters indicates that author-defined boundaries are too variable
to serve as reliable units for downstream QA. The advantage over
semantic chunks and TextTiling segments further shows that topical
coherence alone is insufficient for long-form narrative QA: these
methods do not target the boundaries of self-contained  events
and can therefore produce segments misaligned with the event structure
of long-form narratives.

\begin{table}[t]
  \centering
    \setlength{\abovecaptionskip}{2pt}
  \setlength{\belowcaptionskip}{-3pt}
  \resizebox{\columnwidth}{!}{%
  \begin{tabular}{@{}l@{~~}c@{~~}c@{~~}c@{}}
  \toprule
     \rowcolor{headerblue}
  \textbf{Basic Unit} & \textbf{DetectiveQA} & \textbf{NovelQA} & \textbf{LiteraryQA} \\
  \midrule
  Semantic chunks     & 31.54 & 44.85 & 57.42 \\
  TextTiling segments & 30.54 & 44.60 & 57.01 \\
  Chapters            & 30.37 & 46.07 & 55.81 \\   \rowcolor{bestgold}
  Scenes              & \textbf{35.91} & \textbf{51.45} & \textbf{59.51} \\\bottomrule
  \end{tabular}
  }
  \caption{Effect of basic construction unit on QA accuracy (30B-A3B backbone, top-down pipeline with adaptive retrieval). Per-round retrieval budget is adjusted by unit type to make the token budget comparable to the 20-scene setting. All variants use the same top-down construction pipeline and adaptive retrieval procedure. \textbf{Bold} marks  the best result in each column.}
\label{tab:ablation-units}
\end{table}

\begin{table}[t]
  \setlength{\abovecaptionskip}{2pt}
  \setlength{\belowcaptionskip}{-3pt}
 \resizebox{\columnwidth}{!}{%
\begin{tabular}{@{}l ccc@{}}
\toprule
\rowcolor{headerblue}
\textbf{Retrieval Method} & \textbf{DetectiveQA} & \textbf{NovelQA} & \textbf{LiteraryQA} \\
\midrule
Static RAG \small($k{=}20$)   & 31.38 & 49.84 & 53.97 \\
Static RAG \small($k{=}100$)  & 29.70 & 49.33 & 53.69 \\
Static RAG \small($k{=}200$)  & 29.19 & 49.75 & 53.83 \\
\cmidrule(lr){1-4}
Collapsed RAG \small($k{=}20$)  & 30.20 & 49.52 & \underline{56.09} \\
Collapsed RAG \small($k{=}100$) & 30.54 & 50.07 & 55.05 \\
Collapsed RAG \small($k{=}200$) & 29.19 & 48.28 & 54.67 \\
\cmidrule(lr){1-4}
Iterative RAG \small($k{=}20$/rd)  & \underline{33.39} & \underline{50.44} & 53.31 \\
\cmidrule(lr){1-4}
\rowcolor{bestgold}
\textbf{Adaptive RAG} \small($k{=}20$/rd) & \textbf{35.91} & \textbf{51.45} & \textbf{59.51} \\
\bottomrule
\end{tabular}
}
\caption{Comparison of retrieval policies under different budgets  (30B-A3B backbone). Static RAG retrieves a fixed set of scenes, while Collapsed RAG
retrieves from a fixed pool containing both scenes and tree nodes; iterative
  methods retrieve $k$ scenes per round, capped at 200 in total.  Underlined values mark the strongest non-adaptive baseline;
  \textbf{bold} marks the best result overall.}
  \label{tab:ablation-combined}
\end{table}

\paragraph{Does the gain come from retrieving more context?}
No. Table~\ref{tab:ablation-combined} shows that static retrieval with larger budgets does not
reliably improve performance, and collapsed retrieval over scenes and
tree nodes together remains weaker than adaptive tree-guided retrieval.
Iterative scene-only retrieval is more competitive---multi-step
evidence gathering is clearly useful---but without the storyline tree
it lacks an explicit structure for deciding where to look next.
The tree is most effective when actively used to guide retrieval,
not when flattened into a larger pool of retrievable units. Appendix~\ref{app:case-study} illustrates this behavior with a representative trace, showing how the model uses the storyline tree to navigate from incomplete initial evidence toward crucial evidence for an accurate answer.

\begin{figure}[t]
\centering
\setlength{\abovecaptionskip}{2pt}
\setlength{\belowcaptionskip}{-3pt}
\hspace*{-4pt}\begin{tikzpicture}[scale=.9]
\begin{groupplot}[
  group style={
    group size=2 by 2,
    horizontal sep=.8cm,
    vertical sep=1cm,
  },
  width=5.0cm, height=3.6cm,
  xtick={1,2,3},
  xticklabels={DetQA, NovelQA, LitQA},
  xticklabel style={font=\scriptsize},
  yticklabel style={font=\scriptsize},
  tick style={thin, color=black!30},
  axis line style={thin, color=black!40},
  every axis title/.style={font=\scriptsize\bfseries, at={(0.5,1.05)}, anchor=south},
  enlarge x limits=0.25,
  ymajorgrids=true,
  grid style={dotted, color=black!20},
]

\nextgroupplot[title={(a) \#Storyline nodes / tree}, ymin=0, ymax=390]
\addplot[mark=*, mark size=2.2pt, bucolor, thick,
  error bars/.cd, y dir=both, y explicit]
  coordinates {(1,111.6)+=(0,36.3)-=(0,36.3)
                (2,191.9)+=(0,163.9)-=(0,163.9)
                (3,139.0)+=(0,97.9)-=(0,97.9)};
\addplot[mark=square*, mark size=2pt, tdcolor, thick,
  error bars/.cd, y dir=both, y explicit]
  coordinates {(1,48.3)+=(0,12.3)-=(0,12.3)
                (2,54.8)+=(0,17.3)-=(0,17.3)
                (3,55.0)+=(0,18.2)-=(0,18.2)};

\nextgroupplot[title={(b) \#Top-level nodes / tree}, ymin=0, ymax=13,
legend style={font=\scriptsize, draw=none, fill=white,
              at={(-.2,-1.7)}, anchor=north,
              legend columns=2, column sep=0.5em},
]
\addplot[mark=*, mark size=2.2pt, bucolor, thick,
  error bars/.cd, y dir=both, y explicit]
  coordinates {(1,4.5)+=(0,0.9)-=(0,0.9)
                (2,4.6)+=(0,1.1)-=(0,1.1)
                (3,4.5)+=(0,1.1)-=(0,1.1)};
\addlegendentry{Bottom-up}
\addplot[mark=square*, mark size=2pt, tdcolor, thick,
  error bars/.cd, y dir=both, y explicit]
  coordinates {(1,6.9)+=(0,2.3)-=(0,2.3)
                (2,6.9)+=(0,2.2)-=(0,2.2)
                (3,6.7)+=(0,3.4)-=(0,3.4)};
\addlegendentry{Top-down}

\nextgroupplot[title={(c) \#Bottom-level nodes / tree}, ymin=0, ymax=250]
\addplot[mark=*, mark size=2.2pt, bucolor, thick,
  error bars/.cd, y dir=both, y explicit]
  coordinates {(1,72.9)+=(0,23.8)-=(0,23.8)
                (2,123.3)+=(0,99.5)-=(0,99.5)
                (3,88.4)+=(0,60.1)-=(0,60.1)};
\addplot[mark=square*, mark size=2pt, tdcolor, thick,
  error bars/.cd, y dir=both, y explicit]
  coordinates {(1,13.9)+=(0,4.3)-=(0,4.3)
                (2,16.9)+=(0,7.3)-=(0,7.3)
                (3,16.4)+=(0,6.8)-=(0,6.8)};

\nextgroupplot[title={(d) Avg.\ branching}, ymin=0.6, ymax=3.5,
  ytick={1,2,3}]
\addplot[mark=*, mark size=2.2pt, bucolor, thick,
  error bars/.cd, y dir=both, y explicit]
  coordinates {(1,2.8)+=(0,0.2)-=(0,0.2)
                (2,2.8)+=(0,0.3)-=(0,0.3)
                (3,2.7)+=(0,0.3)-=(0,0.3)};
\addplot[mark=square*, mark size=2pt, tdcolor, thick,
  error bars/.cd, y dir=both, y explicit]
  coordinates {(1,1.5)+=(0,0.5)-=(0,0.5)
                (2,1.6)+=(0,0.5)-=(0,0.5)
                (3,1.7)+=(0,0.7)-=(0,0.7)};

\end{groupplot}
\end{tikzpicture}
\vspace{-0.2em}
\caption{Structural comparison of bottom-up and top-down storyline trees (mean~$\pm$~std
across books). Bottom-up trees are larger overall, with 2--4$\times$ more nodes
concentrated at the bottom level and higher average branching. Top-down trees are
more compact but expose a wider top-level structure, providing a stronger global
narrative prior for retrieval.}
\label{fig:topology-stats}
\begin{subcaptiongroup}
\phantomsubcaption\label{fig:topology-suba}
\phantomsubcaption\label{fig:topology-subb}
\phantomsubcaption\label{fig:topology-subc}
\phantomsubcaption\label{fig:topology-subd}
\end{subcaptiongroup}
\vspace{-0.8em}
\end{figure}
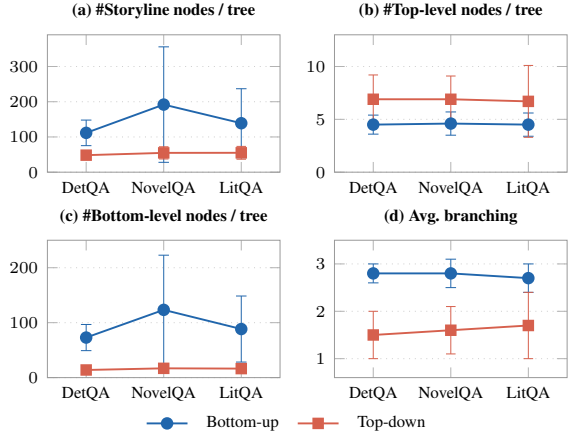

\paragraph{Which trees support more effective retrieval?}
Figure~\ref{fig:topology-suba} shows that bottom-up trees contain
roughly 2--4$\times$ more storyline nodes than top-down trees, with
the excess concentrated at the bottom level immediately above scenes (see Figure~\ref{fig:topology-subc}) 
and driven by consistently higher average branching (see Figure~\ref{fig:topology-subd}). Because bottom-up
clustering starts from scenes, it preserves fine-grained local
groupings but may distribute plotlines across many
local clusters, making the retrieval search space larger
and less aligned with global narrative structure. Top-down construction
instead imposes a compact set of global storylines first and refines
them downward, yielding a more compressed tree with a stronger
global narrative prior that makes it easier for the model to navigate
from broad plotlines to relevant scene-level evidence.

\paragraph{Are top-down trees genuinely more abstract at the top?}
Yes. Despite having fewer nodes overall, top-down trees expose a
\emph{wider} top-level structure than bottom-up trees
(see Figure~\ref{fig:topology-subb}). This means the model is
presented with more distinct high-level plotlines to choose from at
the first retrieval step, enabling coarser but broader navigation
before committing to a specific branch. Bottom-up trees, by contrast,
converge to fewer, more generic top-level clusters that offer less
discriminative entry points for query-guided traversal. Full topology statistics and further analysis are  in Appendix~\ref{app:full-topology-statistics}.

\paragraph{Which question types benefit most from storyline-guided retrieval?}
Table~\ref{tab:novelqa-type-analysis} breaks down NovelQA performance
by question type. Top-down adaptive retrieval yields the largest gains
over TOA on single-hop, detail, plot, character, meaning, and setting
questions --- categories that require locating narrative events and
interpreting them within broader story context, precisely where the
compact global structure of top-down trees is most useful.
Multi-hop performance is nearly tied with TOA, reflecting the
heterogeneity of that category: some multi-hop questions benefit from
storyline-guided exploration, while others require exhaustive coverage
or precise boundary identification rather than selective navigation.
Relation questions are similarly resistant, likely because pairwise
character interactions depend on subtle local evidence in specific
scenes rather than global plotline structure.

\begin{table}[t]
  \centering
  \setlength{\abovecaptionskip}{2pt}
  \setlength{\belowcaptionskip}{-3pt}
  \fontsize{8.5pt}{10pt}\selectfont
  \setlength{\tabcolsep}{3.6pt}
  \renewcommand{\arraystretch}{1.08}
  \begin{tabular}{clrrrr}
  \toprule
     \rowcolor{headerblue}
  \textbf{Dimension} & \textbf{Type} & \multicolumn{1}{c}{\textbf{\%}} & \multicolumn{1}{c}{\textbf{TOA}} & \multicolumn{1}{c}{\textbf{Top-down}} & \multicolumn{1}{c}{\textbf{$\Delta$}} \\
  \midrule
  \multirow{3}{*}{Complexity}
  & Single-hop & 42.74 & 57.11 & \textbf{66.29} & \textbf{+9.18} \\
  & Multi-hop  & 34.98 & \textbf{30.71} & 30.47 & -0.24 \\
  & Detail     & 22.19 & 48.98 & \textbf{59.18} & \textbf{+10.20} \\
  \midrule
  \multirow{7}{*}{Aspect}
  & Plot      & 25.62 & 57.80 & \textbf{67.34} & \textbf{+9.55} \\
  & Times     & 20.07 & \textbf{23.74} & 22.35 & -1.39 \\
  & Meaning   & 15.86 & 44.66 & \textbf{54.08} & \textbf{+9.42} \\
  & Span      & 1.47  & 9.38  & \textbf{9.68}  & +0.30 \\
  & Setting   & 11.44 & 56.12 & \textbf{66.52} & \textbf{+10.41} \\
  & Relation  & 7.15  & \textbf{42.77} & 38.36 & -4.40 \\
  & Character & 18.29 & 53.18 & \textbf{62.15} & \textbf{+8.97} \\
  \bottomrule
  \end{tabular}
  \caption{Question-type analysis on NovelQA using the 30B-A3B backbone. We compare the accuracy of top-down storyline-tree retrieval against TOA across different question types. The \emph{\%} column reports the proportion of questions belonging to each type, and $\Delta$ denotes the accuracy difference for each question type. An explanation of the taxonomy is provided in Appendix \ref{app:novelqa-taxonomy}.}
  \label{tab:novelqa-type-analysis}
  \vspace{-0.8em}
  \end{table}

\section{Conclusion}
\label{sec:conclusion}

We presented storyline trees, a hierarchical representation that
organizes long-form narratives from global themes and major plotlines
down to fine-grained events. Built on scenes as narrative-aware units,
storyline trees provide a reusable representation that bridges broad
narrative organization with concrete evidence. We introduced
both top-down and bottom-up tree induction methods, and used the
resulting trees as an interface for adaptive retrieval in question
answering.

Experiments on three long-context narrative QA benchmarks show that
storyline-tree retrieval consistently outperforms long-context,
memory-based, agentic, and retrieval baselines. Ablations confirm that
these gains come from both scene-based representation and adaptive
tree-guided retrieval, rather than from retrieving more context alone.
Further analysis suggests that top-down storyline trees are especially
effective because their compact global structure helps models navigate
from broad plotlines to relevant scene-level evidence.

Overall, our results show that making narrative structure explicit
offers a  path toward long-context understanding beyond
simply extending context windows. Future work can build on storyline
trees by learning retrieval and traversal policies, and extending
the representation to  narrative understanding tasks beyond
single-book question answering.


\section{Limitations}
\label{sec:limitation}


\paragraph{Prompt-Based Construction and Retrieval}
Our current system is entirely prompt-based. Scene extraction, storyline-tree construction, and adaptive retrieval decisions are all performed through prompting rather than through models explicitly trained to use narrative structure. This makes the approach flexible and easy to apply to new books, but it also limits how reliably the model can exploit the tree: traversal decisions may depend on prompt sensitivity, and the model is not optimized to select the most useful nodes under a retrieval budget. A natural direction for future work is to train models or policies that operate directly over storyline trees, for example by learning which branches to expand, when to stop retrieving, and how to integrate evidence across different levels of abstraction.

\paragraph{Adaptation to Broader Narrative Forms}
Our experiments focus on novels and narrative QA, leaving open how well storyline trees generalize to other long-form narrative genres. Many narrative works do not follow the same prose structure as novels, including long narrative poems, verse novels, movie screenplays, etc. These forms may organize events through stanza structure, dialogue, performance cues, episodes, or recurring symbolic patterns rather than conventional chapters and scenes. Extending storyline trees to these genres may require adapting the definition of basic narrative units and the criteria used to infer higher-level storylines.

\paragraph{Evaluation of Induced Narrative Structure}
We evaluate storyline trees primarily through downstream question answering. While QA provides a direct test of whether the representation helps models locate and use evidence, it does not fully assess the quality of the induced narrative structure itself. For example, a tree may improve retrieval accuracy while still merging distinct plotlines, missing subtle character arcs, or assigning scenes to overly broad storylines. Future work could therefore include intrinsic evaluation of tree faithfulness and interpretability, such as human judgments of scene boundaries, parent-child storyline relations, and whether the hierarchy captures the major narrative arcs of a work.

\paragraph{Computational Cost}
Constructing storyline trees introduces non-trivial overhead relative to single-pass retrieval baselines. Scene extraction requires one LLM inference pass per chapter  to identify scene boundaries and generate structured scene descriptions. Tree construction adds further cost: top-down induction requires  recursive LLM calls to infer and refine storyline nodes, while bottom-up construction requires embedding, clustering, and cluster summarization at each level. This overhead is partly amortized when a tree is reused across multiple queries on the same book,  but may limit scalability to very large collections or latency-sensitive indexing  settings. Future work could reduce this overhead by employing smaller or distilled models for scene extraction, caching reusable narrative structures, constructing trees incrementally, or learning lightweight traversal policies that selectively expand only subtrees relevant to a given query or task. 

\bibliography{custom}

\newpage

\appendix


\begin{table}[t]
\begin{tracefloatbox}{Example Scene from Wuthering Heights}
\textbf{Title:} \texttt{Zillah Rescues the Narrator from Heathcliff's Household}\\[2pt]
\textbf{Synopsis:} \texttt{Zillah intervenes, rescues the narrator from the dogs, treats his injuries, and arranges for him to be taken to bed...}\\[2pt]
\textbf{Significance:} \texttt{Introduces Zillah as a rare figure of compassion and rationality...}\\[2pt]
\textbf{End Sentence:} \texttt{...ushered me to bed.}
\end{tracefloatbox}
\caption{Example output schema for scene extraction. \emph{End Sentence} is shown here in text form for readability; in implementation, it is represented by the corresponding sentence index to enable stable alignment with the source text, as specified in the scene-extraction prompt in Section~\ref{app:prompt-template}.}
\label{tab:scene-fields}
\end{table}

\section{Experiment Details}
\label{app:exp-details}


\subsection{Datasets}
\label{app:dataset-stats}

For each book, we use the corresponding source text as the input
document. Since our scene extraction pipeline processes books at the
chapter level, we first attempt to recover chapter boundaries from
metadata and structural markers already present in the source text.
When such information is missing or unreliable, we infer chapter
boundaries from ebook structure, title-like headings, and surrounding
content, producing an ordered chapter sequence while preserving the
original narrative order. All source-text preprocessing and
storyline-tree construction are performed without access to the
questions or gold answers. The average input length in
Table~\ref{tab:dataset_stats} is computed over the full preprocessed
source text using the tokenizer of the backbone model used in our
experiments.

\begin{table*}[t]
  \centering
  \small
  \setlength{\tabcolsep}{2.5pt}
  \renewcommand{\arraystretch}{1.08}
  \begin{tabular}{@{}>{\raggedright\arraybackslash}p{0.4\columnwidth}
                  >{\raggedright\arraybackslash}p{0.42\columnwidth}
                  >{\raggedright\arraybackslash}p{0.42\columnwidth}
                  >{\raggedright\arraybackslash}p{0.42\columnwidth}@{}}
  \toprule
   & \multicolumn{1}{c}{\textbf{DetectiveQA}}
   & \multicolumn{1}{c}{\textbf{NovelQA}}
   & \multicolumn{1}{c}{\textbf{LiteraryQA}} \\
  \midrule
  \# books/docs
  & \multicolumn{1}{c}{62}
  & \multicolumn{1}{c}{85}
  & \multicolumn{1}{c}{105} \\
  \# QA pairs
  & \multicolumn{1}{c}{596}
  & \multicolumn{1}{c}{2,177}
  & \multicolumn{1}{c}{2,872} \\
  \# Avg. input tokens
  & \multicolumn{1}{c}{87,588}
  & \multicolumn{1}{c}{186,669}
  & \multicolumn{1}{c}{115,083} \\
  \midrule
  \emph{Primary focus}
  & Detective narratives requiring clue synthesis and implicit inference.
  & Evidence-grounded QA spanning single-hop, multi-hop, and fine-grained detail questions.
  & Open-ended QA emphasizing cross-passage synthesis and broader narrative understanding. \\
  \bottomrule
  \end{tabular}
  \caption{Statistics and primary focus of the narrative QA datasets used in our experiments. We report the number of books or documents, the number of question-answer pairs, and the average input length in tokens.}
  \label{tab:dataset_stats}
\end{table*}

\subsection{Baseline and Variant Implementation Details}
\label{app:baseline-details}

This section provides implementation details for comparison methods and ablation variants that involve external implementations or non-obvious design choices. Methods whose implementation is directly determined by the definitions in Section~\ref{sec:experiments} are summarized only through the shared protocol: zero-shot directly answers from the preprocessed source text, truncated when necessary; chapter-based construction replaces scenes with recovered chapters as the basic units; and retrieval baselines use the same dense retriever and answer-generation format, differing only in the evidence source or retrieval policy. Unless otherwise stated, all methods use the same generative answer format as the zero-shot QA prompt. Retrieval-based variants use Qwen3-Embedding-8B as the dense retriever~\citep{zhang2025qwen3embedding} and follow the same evidence formatting as our method.

\subsubsection{Long-Context and Agentic Baselines}

\paragraph{\textsc{QwenLong-L1.5}}
\textsc{QwenLong-L1.5} is a post-training recipe for improving long-context reasoning and memory management~\citep{shen2025qwenlongl15}. The method constructs long-context training data by decomposing documents into atomic facts and composing questions whose answers require integrating evidence distributed across the input. It further employs reinforcement-learning with mechanisms for balancing long-context tasks and controlling the model's exploration during policy optimization. In our experiments, we evaluate the released \textsc{QwenLong-L1.5} checkpoint as a long-context post-trained model, rather than reimplementing its training recipe. When the preprocessed source text fits within the model context window, we use the same generative QA setting as the zero-shot baseline and directly prompt the model with the source text and question. When the source text exceeds the context window, we follow the memory-management behavior provided by the original \textsc{QwenLong-L1.5} implementation: the document is processed in long-context segments, intermediate information is maintained through the model's memory mechanism, and the final answer is generated from the question together with the accumulated memory state. Since no 235B-A22B checkpoint is available, we report this baseline only in the 30B-A3B setting.

\paragraph{\textsc{MemAgent}}
\textsc{MemAgent} formulates long-document processing as a sequential interaction between a language model and an auxiliary memory state~\citep{yu2025memagent}. The document is read segment by segment; after each segment, the agent updates a compact memory that is carried forward to the next step. This design allows the model to process arbitrarily long inputs by retaining selected information from earlier segments while discarding less useful details. The original method trains the memory-update behavior with reinforcement learning so that the agent learns how to write, overwrite, and preserve information across a long sequence. In our experiments, we follow the original inference procedure and prompting scheme. The input is processed as generic text segments, the final answer is generated from the question and accumulated memory.

\paragraph{\textsc{Tree of Agents}}
\textsc{Tree of Agents} (TOA) decomposes a long input into chunks and assigns chunk-level agents to process different parts of the document~\citep{yu2025treeagents}. Each agent first reads its assigned chunk and produces a local cognitive state consisting of supporting evidence and a tentative answer. These cognitive states are stored in agent-specific buffers and then shared across agents. In the multi-perspective reasoning phase, each agent inspects the cognitive states produced by other agents and decides which additional chunks may help answer the question. TOA then probes different orders of these selected chunks as tree-structured paths: along each path, the agent incrementally reads additional chunks and updates its cognitive state. To reduce redundant computation, TOA caches intermediate states that share the same path prefix and prunes paths whose newly read chunks are judged unhelpful. In the final consensus phase, each agent may have multiple path-specific states; the method selects the longest cached chunk sequence for that agent and uses it to produce a local final answer. The overall answer is then obtained by majority voting over the local final answers from all agents. In our experiments, we follow the original agent organization and prompting scheme.

\subsubsection{Tree-Construction Variants}

\paragraph{\textsc{RAPTOR}} constructs a retrieval tree by recursively clustering and summarizing text units~\citep{sarthi2024raptor}. Starting from leaf chunks, the method embeds the current units, reduces the embedding space, clusters related units, and summarizes each cluster into a higher-level node. The newly generated summaries are then treated as units for the next recursion step, yielding a hierarchy in which lower levels retain fine-grained information and upper levels provide increasingly abstract summaries. Retrieval can then access information at multiple levels of this hierarchy rather than only from the original chunks. Our main bottom-up variant follows this recursive cluster-and-summarize framework, but adapts the leaves and summaries to our narrative setting. We start from extracted scenes rather than fixed-size chunks, represent scene leaves using their structured scene fields, and summarize each cluster into a storyline node following the title--description schema used throughout our paper. Clustering follows RAPTOR: we apply UMAP dimensionality reduction followed by a Gaussian mixture model, selecting the number of clusters using the Bayesian information criterion (BIC)~\citep{schwarz1978estimating}. The recursive procedure continues until the current level can no longer be meaningfully clustered.

\paragraph{\textsc{G-RAPTOR}} extends the RAPTOR-style hierarchy by replacing the standard clustering step with graph-based community detection~\citep{liu2026semanticraptor}. In the original formulation, the method first builds a similarity graph over embedded units and then applies an adaptive graph-clustering algorithm to form higher-level groups. This changes the topology of the induced hierarchy: rather than relying on mixture-model clusters in a reduced embedding space, it groups units according to graph connectivity and community structure. In our implementation, we keep the same recursive bottom-up summarization framework and the same scene-based leaf units but replace the GMM-based clustering step with graph clustering. Specifically, we construct a cosine-similarity nearest-neighbor graph over the current node embeddings and apply Leiden clustering to form groups for summarization. Each resulting group is summarized into a storyline node using the same title--description schema, so this variant differs from our \textsc{RAPTOR} implementation only in the grouping step.

\paragraph{\textsc{N-RAPTOR}}  is an order-aware bottom-up variant specific to our experiments. It follows the same recursive cluster-and-summarize framework as the other bottom-up variants, but replaces global semantic clustering with a greedy contiguous grouping procedure. At each level, the method scans the ordered sequence of units from left to right, maintaining a current group and deciding whether the next adjacent unit should be merged into it based on its semantic compatibility with both the current group and the most recent unit. If the next unit is not sufficiently compatible, the current group is closed and a new group is started. This produces contiguous clusters that respect the narrative order while still using embedding similarity to decide local group boundaries. Each group is then summarized into a storyline node with the same title--description schema, and the resulting nodes are recursively grouped and summarized until no further reduction is possible.

\subsubsection{Segmentation Variants}
\label{app:segmentation-variants}

For the basic-unit ablation in Table~\ref{tab:ablation-units}, different segmentation methods produce units with substantially different average lengths. To control for this variation, we adjust the per-round retrieval budget for each unit type so that the average
amount of retrieved source text per round is approximately equal to the 20-scene setting. Table~\ref{tab:unit-token-budget} reports the
resulting budgets; since unit lengths vary across books and budgets must be integer-valued, the token match is approximate.

\begin{table}[t]
  \centering
  \small
  \setlength{\tabcolsep}{5pt}
  \renewcommand{\arraystretch}{1.05}
  \begin{tabular}{@{}lcc@{}}
    \toprule
    \rowcolor{headerblue}
    \textbf{Basic Unit} & \textbf{Avg. Tokens / Unit} & \textbf{Budget / Round} \\
    \midrule
    Semantic chunks     & 516  & 18 \\
    TextTiling segments & 165  & 60 \\
    Chapters            & 3841 & 3  \\
    Scenes              & 465  & 20 \\
    \bottomrule
  \end{tabular}
  \caption{Average unit length and per-round retrieval budget for each
  segmentation method in the basic-unit ablation
  (Table~\ref{tab:ablation-units}). Budgets are set so that each
  method retrieves approximately the same number of source tokens
  per round as the 20-scene setting, making accuracy differences
  attributable to segmentation quality rather than context volume.}
  \label{tab:unit-token-budget}
\end{table}

\paragraph{TextTiling segments}
TextTiling is an unsupervised discourse segmentation algorithm that identifies subtopic boundaries through lexical cohesion~\citep{hearst1997texttiling}. The method compares adjacent blocks of text and detects boundaries at points where lexical similarity drops, producing contiguous multi-paragraph segments that correspond to shifts in local topic.  We use TextTiling  as a generic segmentation baseline for replacing scenes as the basic units. The resulting segments are treated as leaf units in the same top-down construction pipeline used for scenes.

\paragraph{Semantic Chunks}
Semantic chunking segments a document according to semantic coherence rather than fixed length or explicit chapter boundaries~\citep{qu2024semanticchunking}. A common breakpoint-based implementation embeds consecutive text spans and inserts a boundary when the semantic distance between neighboring spans indicates a topic shift. This produces contiguous chunks intended to align better with changes in document content than fixed-size windows. In our experiments, we use breakpoint-based semantic chunks as an alternative set of leaf units for top-down construction, where the chunks replace scenes as the basic units.

\begin{table*}[t]
  \centering
  \small
  \setlength{\tabcolsep}{4.5pt}
  \begin{tabular}{cccrccccc}
    \toprule
    \textbf{Backbone} & \textbf{Dataset} & \textbf{Baseline}
    & \multicolumn{1}{c}{\textbf{\#Q}} & \textbf{Top-down Acc.} & \textbf{Baseline Acc.}
    & \textbf{$\Delta$} & \textbf{95\% CI} & \textbf{$p$} \\
    \midrule
    30B-A3B   & DetectiveQA & TOA       & 596  & 35.91 & 32.72 & +3.19 & $[-0.67, +7.05]$ & 0.056 \\
    30B-A3B   & NovelQA     & Zero-shot & 2177 & 51.45 & 48.78 & +2.66 & $[+0.64, +4.69]$ & 0.005 \\
    30B-A3B   & LiteraryQA  & TOA       & 2872 & 59.51 & 58.25 & +1.25 & $[-0.56, +3.06]$ & 0.093 \\
    235B-A22B & DetectiveQA & TOA       & 596  & 40.27 & 38.76 & +1.51 & $[-2.35, +5.37]$ & 0.233 \\
    235B-A22B & NovelQA     & TOA       & 2177 & 60.27 & 58.06 & +2.20 & $[+0.32, +4.09]$ & 0.012 \\
    235B-A22B & LiteraryQA  & TOA       & 2872 & 69.19 & 68.14 & +1.04 & $[-0.63, +2.72]$ & 0.116 \\
    \bottomrule
  \end{tabular}
  \caption{Question-level paired bootstrap significance tests for the main results. Accuracies, gains, and confidence intervals are reported in percentage points. For each dataset--backbone setting, the comparator is the strongest non-storyline baseline in Table~\ref{tab:qa-results}. We use 100{,}000 bootstrap resamples and report one-sided $p$-values for the directional hypothesis that top-down storyline-tree retrieval improves over the comparator.}
  \label{tab:main-significance}
\end{table*}

\subsection{NovelQA Question Taxonomy}
\label{app:novelqa-taxonomy}

NovelQA provides two annotations for each question: a complexity label
and an aspect label \citep{wang2025novelqa}. We use these labels only for diagnostic analysis in Table~\ref{tab:novelqa-type-analysis}; they are not provided to the
model during retrieval or answer generation.

\paragraph{Complexity Labels}
The complexity dimension describes how evidence is distributed in the
novel and how much integration is required to answer the question.
\begin{description}
    \item[Multi-hop] questions whose answer requires combining
    information from multiple non-adjacent passages, scenes, or
    chapters. These questions test whether a model can retrieve and
    integrate evidence that is distributed across the narrative.

    \item[Single-hop] questions that can be answered from a localized
    passage or a small contiguous context. They still require finding
    the relevant part of the novel, but the answer does not depend on
    combining widely separated pieces of evidence.

    \item[Detail] questions that ask about minor, low-salience
    information, usually grounded in a very small local context. These
    differ from ordinary single-hop questions because the relevant fact
    is often peripheral to the main plot and therefore difficult to
    recover from coarse summaries or high-level memory.
\end{description}

\paragraph{Aspect Labels}
The aspect dimension describes the type of narrative information
targeted by the question.
\begin{description}
    \item[Times] questions about frequency, such as how often a
    character, event, location, interaction, or plot pattern appears in
    the novel.

    \item[Meaning] questions that require interpreting narrative
    meaning, including implications, symbols, metaphors, paraphrased
    sentences, or episodes that signal a change in character or theme.

    \item[Span] questions about the full temporal or geographical
    range of the story, such as the earliest and latest years covered
    by the narrative or the complete set of places involved.

    \item[Setting] questions about time or place settings that do not
    require recovering the full range of the narrative. These typically
    concern where or when a specific event or episode takes place.

    \item[Relation] questions about relations involving multiple
    character entities, including interpersonal relations, aliases,
    designations, or group membership.

    \item[Character] questions centered on character information that
    is not primarily relational, such as identifying a character,
    describing their role, or retrieving character-specific facts.

    \item[Plot] questions about narrative events and actions, including
    what happens, whether an event occurs, what a character does, or how
    an episode develops.
\end{description}

\begin{table*}[t]
  \centering
  \small
  \setlength{\tabcolsep}{4.2pt}
  \renewcommand{\arraystretch}{1.08}
  \begin{tabular*}{\textwidth}{@{\extracolsep{\fill}}lcccccc@{}}
  \toprule
  \multirow{2}{*}{\textbf{Metric}}
  & \multicolumn{2}{c}{\textbf{DetectiveQA}}
  & \multicolumn{2}{c}{\textbf{NovelQA}}
  & \multicolumn{2}{c}{\textbf{LiteraryQA}} \\
  \cmidrule(lr){2-3} \cmidrule(lr){4-5} \cmidrule(lr){6-7}
  & \textbf{Bottom-up} & \textbf{Top-down}
  & \textbf{Bottom-up} & \textbf{Top-down}
  & \textbf{Bottom-up} & \textbf{Top-down} \\
  \midrule

  Depth
  & 3.9 $\pm$ 0.4 & 4.0
  & 4.2 $\pm$ 0.7 & 4.0
  & 4.0 $\pm$ 0.7 & 4.0 \\

  \#Storyline nodes / tree
  & 111.6 $\pm$ 36.3 & 48.3 $\pm$ 12.3
  & 191.9 $\pm$ 163.9 & 54.8 $\pm$ 17.3
  & 139.0 $\pm$ 97.9 & 55.0 $\pm$ 18.2 \\

  \#Top-level nodes / tree
  & 4.5 $\pm$ 0.9 & 6.9 $\pm$ 2.3
  & 4.6 $\pm$ 1.1 & 6.9 $\pm$ 2.2
  & 4.5 $\pm$ 1.1 & 6.7 $\pm$ 3.4 \\

  \#Bottom-level nodes / tree
  & 72.9 $\pm$ 23.8 & 13.9 $\pm$ 4.3
  & 123.3 $\pm$ 99.5 & 16.9 $\pm$ 7.3
  & 88.4 $\pm$ 60.1 & 16.4 $\pm$ 6.8 \\

  Avg. branching
  & 2.8 $\pm$ 0.2 & 1.5 $\pm$ 0.5
  & 2.8 $\pm$ 0.3 & 1.6 $\pm$ 0.5
  & 2.7 $\pm$ 0.3 & 1.7 $\pm$ 0.7 \\

  Max branching
  & 5.6 $\pm$ 0.7 & 5.9 $\pm$ 3.7
  & 5.8 $\pm$ 0.8 & 5.6 $\pm$ 3.1
  & 5.6 $\pm$ 1.1 & 7.0 $\pm$ 4.7 \\

  \bottomrule
  \end{tabular*}
  \caption{Topology statistics of storyline trees, reported as mean
  $\pm$ standard deviation across books. We count only storyline nodes
  and exclude scene leaves. Depth denotes the number of storyline
  levels. Top-level nodes/tree and bottom-level nodes/tree refer to the
  most abstract storyline level and the most specific storyline level
  immediately above scenes, respectively. Avg. branching and Max
  branching are computed over the storyline-node hierarchy.}
  \label{tab:topology-stats}
\end{table*}

\section{Significance Testing}
\label{app:significance}

We assess the reliability of the main QA gains using a paired bootstrap test. For each dataset-backbone setting in Table~\ref{tab:qa-results}, we compare the top-down storyline-tree method against the strongest non-storyline comparator in that setting. Let $z_i^{\mathrm{TD}}$ and $z_i^{\mathrm{base}}$ denote the judged correctness of the two systems on question $i$, where each value is binary. The observed accuracy difference is
\begin{equation}
\Delta = \frac{100}{N}\sum_{i=1}^{N}
\big(z_i^{\mathrm{TD}} - z_i^{\mathrm{base}}\big).
\end{equation}
We draw 100{,}000 bootstrap resamples of questions with replacement and recompute $\Delta$ for each resample. We report the percentile 95\% confidence interval of the resulting bootstrap distribution, together with a one-sided bootstrap $p$-value for the directional hypothesis that top-down storyline-tree retrieval improves over the comparator. Missing predictions are counted as incorrect.

Table~\ref{tab:main-significance} shows that top-down storyline-tree retrieval obtains positive gains over the strongest non-storyline comparator in all six main comparisons. The strongest statistical support appears on NovelQA, where the gains are significant for both the 30B-A3B and 235B-A22B backbones, indicating that the improvement is robust across model scales. DetectiveQA and LiteraryQA also show consistent positive gains, although their 95\% confidence intervals include zero under question-level bootstrap. We therefore interpret these results as directional rather than conclusive evidence, while noting that the overall pattern is consistently favorable to storyline-tree retrieval.

\section{Full Topology Statistics}
\label{app:full-topology-statistics}

Table~\ref{tab:topology-stats} reports the complete topology
statistics underlying the structural analysis in
Figure~\ref{fig:topology-stats}. In addition to the quantities shown in
the main text, the table includes tree depth and maximum branching.
The results show that top-down and bottom-up trees have comparable
depths, while differing mainly in how nodes are distributed across
levels. Bottom-up construction produces substantially more storyline
nodes, especially at the lowest storyline level above scenes, whereas
top-down construction yields a more compact hierarchy with a broader
top level. Maximum branching is broadly comparable across the two
methods, suggesting that the main structural difference lies in overall
tree size and level-wise allocation rather than occasional extreme
branching.

\section{Bottom-up Variants}
\label{app:bottom-up-variants}

We further compare three bottom-up variants that share the same
cluster-and-summarize framework but differ in how lower-level units are
grouped. \textsc{RAPTOR} denotes our main bottom-up construction, which
follows the recursive clustering procedure of \citet{sarthi2024raptor}
and summarizes each cluster into a storyline node using the
title-description schema in Section~\ref{sec:tree-construction}.
\textsc{G-RAPTOR} replaces the GMM-based clustering step with graph-based
community detection: it builds a cosine-weighted $k$-nearest-neighbor
graph over node embeddings and applies Leiden clustering, following
\citet{liu2026semanticraptor}. \textsc{N-RAPTOR} is an order-aware
variant that only groups adjacent units in the narrative sequence,
thereby preserving local plot continuity during recursive aggregation.

Table~\ref{tab:bottom-up-methods} shows that no single alternative
dominates across all datasets. The standard \textsc{RAPTOR} variant
performs best on DetectiveQA and LiteraryQA, while the order-aware
\textsc{N-RAPTOR} variant achieves the highest score on NovelQA. This
suggests that different clustering assumptions capture different aspects
of narrative structure: semantic clustering can connect thematically
related scenes even when they are far apart, whereas adjacency-constrained
clustering better preserves local plot continuity. Overall, we use
\textsc{RAPTOR} as the default bottom-up construction in the main
experiments because it gives the strongest average performance and is
more robust across datasets.

\section{Case Study: Adaptive Retrieval Trace}
\label{app:case-study}

Table~\ref{tab:case-study-trace} illustrates how adaptive retrieval
progressively focuses from broad narrative context toward the precise
scene-level evidence needed to answer a question. Given the question
\textit{``Why were Petrina and Irimiás summoned?''}, the model first
receives scenes retrieved directly from the question. These scenes
establish that Petrina and Irimiás are waiting in a government office
to respond to an official accusation, but leave the cause of
the summons unspecified. The model next inspects the broad storyline \emph{Irimiás and Petrina's
Return and Reestablishment of Control}, which retrieves scenes from the
larger arc of their return and subsequent actions. This confirms the
relevant narrative arc but still does not resolve the accusation. The
model then selects the more focused storyline \emph{Irimiás and Petrina
face bureaucratic humiliation in the government office} which 
surfaces the decisive evidence: in \emph{Misunderstanding
the Summons}, Petrina declares that their job is to supply information,
and in \emph{The Captain's Ultimatum}, the captain says that they were
summoned because they endangered the project by their absence. The trace
therefore demonstrates how the storyline tree acts as an interpretable navigation map from
global plot context to the local scene-level evidence that grounds the
final answer.

\begin{table}[t!]
  \centering
  \small
  \setlength{\tabcolsep}{4pt}
  \renewcommand{\arraystretch}{1.05}
  \begin{tabular}{@{}lccc@{}}
    \toprule
    \textbf{Method} & \textbf{DetectiveQA} & \textbf{NovelQA} & \textbf{LiteraryQA} \\
    \midrule
    \textsc{RAPTOR}   & \textbf{34.06} & 50.99 & \textbf{58.64} \\
    \textsc{G-RAPTOR} & 33.89 & 50.07 & 57.83 \\
    \textsc{N-RAPTOR} & 33.72 & \textbf{51.72} & 56.20 \\
    \bottomrule
  \end{tabular}
  \caption{Comparison of bottom-up tree-construction variants using the
  Qwen3-30B-A3B-Instruct-2507 backbone. All variants use the same
  adaptive retrieval procedure; they differ only in the clustering step
  used during bottom-up tree construction.}
  \label{tab:bottom-up-methods}
\end{table}


\begin{tracebox}[fontupper=\fontsize{8.3pt}{9.4pt}\selectfont]{Case Study: Adaptive Retrieval Trace}

\textbf{Question:} Why were Petrina and Irimiás summoned?\\
\textbf{Reference answer:} Because their absence endangered the project.\\[2pt]

\textbf{Setup:}
The model is given the complete storyline tree and an initial set of
scenes retrieved directly from the question. At each step, it must either
answer or select one storyline title for additional scene retrieval.\\[2pt]

\texttt{<storylines>} \emph{Relevant nodes shown.}\\
\textbf{Level 1 storyline:}
\emph{Irimiás and Petrina's Return and Reestablishment of Control}.\\
Description: Irimiás and Petrina arrive at a government office to face an
unspecified accusation, only to discover they are in the wrong place.
After enduring bureaucratic humiliation and identity confusion, they
leave and travel through a storm toward the estate.\\
\textbf{Level 2 storyline:}
\emph{Irimiás and Petrina face bureaucratic humiliation in the government office}.\\
Description: Irimiás and Petrina arrive at a government building expecting
to answer an accusation, only to discover they are in the wrong location.
Subjected to mockery and procedural confusion, they endure a degrading
encounter with officials.\\
\texttt{</storylines>}\\[2pt]

\texttt{<initial\_scenes>} \emph{Titles only.}\\
Irimiás's Doubt; The Kid's Rebellion; The Clocks and the Poet; The
Journey Begins; Irimiás's Breakdown; Irimiás's Accusation and
Condemnation; Journey to the Estate; Petrina's Plea; Irimiás's
Prophecy; The Corpse Rises; The Mocking Summons; The Call for
Confession; The Horses Seen; God and the Network; Approaching the
Manor; Insult and Interruption; Night's End; The New Plan Revealed;
Waiting in the Corridor; Kelemen Tells of Irimiás.\\
\texttt{</initial\_scenes>}

\par\smallskip\noindent{\color{sceneFieldsColor}\rule{\linewidth}{0.35pt}}\par\smallskip

\textbf{Step 1: Initial evidence is relevant but under-specified.}\\
\textbf{Reasoning:} The initial scenes include multiple references to a
summons received by Irimiás and Petrina, particularly in ``Chapter: We
Are Resurrected, Scene: Waiting in the Corridor,'' where they sit in a
government office waiting to respond to an accusation. The text mentions
a ``formal summons'' with language about ``the shadow of suspicion that
has fallen,'' indicating an official investigation. However, the exact
reason for the summons is not specified in the provided scenes. To
determine the cause, I need to retrieve a storyline that details the
nature of the accusation they faced.\\
\textbf{Decision:} retrieve.\\
\textbf{Storyline Title:} \emph{Irimiás and Petrina's Return and Reestablishment of Control}.

\par\smallskip\noindent{\color{sceneFieldsColor}\rule{\linewidth}{0.35pt}}\par\smallskip

\textbf{Retrieved scenes for requested storyline title:} \emph{Titles only.}\\
The Kid's Rebellion; The Journey Begins; Journey to the Estate;
Irimiás's Prophecy; Irimiás's Doubt; Irimiás's Breakdown; The Network
Revealed; Storming Through Town; Irimiás Arrives; Irimiás's Accusation
and Condemnation; Irimiás's Reprimand; The Clocks and the Poet; Fantasy
of Destruction; The New Plan Revealed; Petrina's Plea; Irimiás's
Departure and Aftermath; Approaching the Manor; Aftermath of the
Vision; The Call for Confession; Insult and Interruption.

\par\smallskip\noindent{\color{sceneFieldsColor}\rule{\linewidth}{0.35pt}}\par\smallskip

\textbf{Step 2: Broad storyline confirms the arc but not the cause.}\\
\textbf{Reasoning:} The initial scenes and the retrieved scenes from
``Irimiás and Petrina's Return and Reestablishment of Control'' contain
multiple references to a summons and their arrival at a government
office, but they do not specify the exact reason for the summons. The
description of the storyline mentions that Irimiás and Petrina face an
``unspecified accusation'' and endure bureaucratic humiliation, but the
nature of the accusation is not detailed. To determine why they were
summoned, I need to retrieve a storyline that explicitly addresses the
accusation or the events leading to the summons.\\
\textbf{Decision:} retrieve.\\
\textbf{Storyline Title:} \emph{Irimiás and Petrina face bureaucratic humiliation in the government office}.

\par\smallskip\noindent{\color{sceneFieldsColor}\rule{\linewidth}{0.35pt}}\par\smallskip

\textbf{Retrieved scenes for requested storyline title:} \emph{Titles only.}\\
Waiting in the Corridor; Entering the Stockroom; Insult and
Interruption; Misunderstanding the Summons; Irimiás's Doubt; The Mocking
Summons; Journey to the Estate; The Kid's Rebellion; Irimiás's
Prophecy; The Clocks and the Poet; Irimiás's Breakdown; Name Confusion;
Storming Through Town; The Captain's Ultimatum; Collapse of Negotiation;
The Journey Begins; The Sergeant's Warning; Fantasy of Destruction; The
Captain Arrives; Irimiás's Reprimand.

\par\smallskip\noindent{\color{sceneFieldsColor}\rule{\linewidth}{0.35pt}}\par\smallskip

\textbf{Step 3: Focused storyline surfaces decisive evidence.}\\
\textbf{Reasoning:} The retrieved scenes from ``Irimiás and Petrina
face bureaucratic humiliation in the government office'' contain
critical information about the summons. In ``Scene: The Captain's
Ultimatum,'' the captain explicitly states that they were summoned
because ``you have endangered the project by your absence.'' He further
clarifies that their role is expected to continue under his authority,
indicating that the summons was not about a criminal accusation but
about their failure to fulfill their duties within a system they
previously served---likely as informants. This is supported by Petrina's
outburst in ``Scene: Misunderstanding the Summons,'' where he declares,
``You know very well our job is to supply information,'' confirming
their role. The accusation is not about a specific crime but about
dereliction of duty in an ongoing, unnamed project.\\
\textbf{Decision:} answer.\\
\textbf{Answer:} They were summoned for endangering a project by their absence.

\end{tracebox}
\promptcaption{A two-round adaptive retrieval trace using the storyline tree.
The initial retrieval identifies relevant but incomplete
evidence; the model then requests a broad storyline before selecting a
more focused storyline that retrieves the decisive scene-level evidence.
Full scenes are abbreviated to titles.}{tab:case-study-trace}

\section{Prompt Templates}
\label{app:prompt-template}

In this section, we provide the prompt templates used in this paper. The zero-shot QA prompt shown in Table~\ref{tab:prompt-zero-shot-qa} is used by the \emph{Zero-shot} baseline. For \textsc{QwenLong-L1.5}, \emph{MemAgent}, and \emph{TOA}, we follow the model- or method-specific prompt templates from their original implementations.

The retrieval ablations, including scene-level RAG and collapsed retrieval, use analogous QA templates that preserve the answer-generation format of the zero-shot prompt, with references to the full novel replaced by retrieved evidence.

The scene-extraction prompt in Table~\ref{tab:prompt-scene-extraction} operates on chapters that are first sentence-segmented with spaCy\footnote{https://spacy.io/} and annotated with sentence numbers; the model outputs sentence-level boundary indices, rather than raw boundary strings, to support stable alignment with the source text. 

The two top-down prompts (Table~\ref{tab:prompt-topdown-first} and \ref{tab:prompt-topdown-subsequent}) first infer major storylines and then recursively decompose them into narrower sub-storylines; in the recursive prompt, parent storylines are numbered and the model returns parent indices to define tree edges reliably. The adaptive-retrieval prompt in Table~\ref{tab:prompt-adaptive-retrieval} initializes the QA agent with the storyline tree and question-based scene evidence, using a strict retrieve-or-answer response format. The evaluation prompt in Table~\ref{tab:prompt-evaluation} is used by the LLM judge to determine whether a model answer is equivalent to an accepted reference answer.

\begin{promptbox}[title={Zero-Shot QA Prompt}]
You are a literature professor. I will provide you with the full text of a novel along with a question. Please thoroughly analyze the novel to accurately respond to the following question.

Book Content:
<text>
{content}
</text>

Question: {question}

Try your best to answer the question based on the given novel full text. The correct answer should be in short with only one or several words. Give a brief explanation for your answer starting with "Explanation:", then output the answer starting with "Answer:" and the actual answer in "\boxed{answer}". For example:

Explanation: In a few sentences, citing evidence from the novel appropriately.
Answer: \boxed{answer in several words}
\end{promptbox}
\captionof{table}{QA prompt for the Zero-shot baseline.}
\label{tab:prompt-zero-shot-qa}


\begin{promptbox}[title={Evaluation Prompt}]
You are grading answer correctness for a QA task.

You are NOT answering the question from the novel.
You are ONLY deciding whether the model answer is semantically equivalent to an acceptable reference answer.

Rules:
1. Use only the question, the reference answer(s), and the model answer.
2. Do not use outside knowledge.
3. Ignore minor wording differences, articles, punctuation, tense, and simple paraphrases.
4. Judge CORRECT only if the model answer clearly refers to the same person, object, event, relation, or cause as at least one acceptable reference answer.
5. Accept concise aliases or descriptive paraphrases only when the equivalence is clear from the texts provided.
6. Judge INCORRECT if the model answer is ambiguous, mentions multiple candidates, is broader or narrower than the reference answer, contradicts it, or does not clearly match it.
7. Do not force a positive match when the evidence is weak.
8. If multiple reference answers are listed, treat them as alternative acceptable gold answers and mark CORRECT if the model answer clearly matches any one of them.

Return JSON only with this schema:
{
  "is_correct": true | false,
  "confidence": "high" | "medium" | "low",
  "reason": "one short sentence"
}

Question:
{question}

Accepted reference answer(s):
{reference_answers}

Model answer:
{answer_text}
\end{promptbox}
\captionof{table}{Prompt used by the LLM judge to score equivalence between a model answer and  reference answer(s).}
\label{tab:prompt-evaluation}


\begin{promptbox}[title={Scene Extraction Prompt}]
You are an expert literary editor and structural analyst. Your task is to identify and extract the *major narrative scenes* from the provided book chapter.

# 1. Definition of a Scene
A **scene** is a distinct unit of dramatic action. While scenes often follow the unity of time and place, **Narrative Unity** is the most important factor.

**Criteria for a Scene:**
* **The Container:** It typically occurs in a continuous block of time.
* **The Content:** It centers on a character pursuing a specific goal or dealing with a specific conflict.
* **The Change:** It results in a change of state (emotional, physical, or informational) by the end.

**Criteria for Scene Breaks (When to Cut):**
* **Significant Time Jump:** The narrative moves from "afternoon" to "evening," or skips days/years.
* **Major Location Shift:** A complete change of setting (e.g., moving from a house to a workplace).
    * *Exception:* Do NOT create a new scene for minor transit (e.g., walking from the kitchen to the living room, or driving a car) unless a major plot twist occurs specifically during the transit. Merge these transitions into the main scene they support.
* **POV Switch:** The narrator changes (e.g., strictly switching from Character A's head to Character B's head).

**What is NOT a Scene:**
* Brief moments of introspection without action.
* Short transitions between locations.
* Summary passages ("The next three months were hard...") -- treat these as the introduction to the next scene or the conclusion of the previous one.

# 2. Strict Constraints (Crucial)
1.  **Full Coverage:** The extracted scenes must cover the **entirety** of the chapter text. There must be no "gaps" or text left out between scenes. `ending_sentence` represents the end of the scene as the sentence number (e.g., `5` means sentence number 5, which corresponds to the sentence labeled `[5]` in the chapter text). The first scene will automatically start at the beginning of the chapter text, and every subsequent scene will start from the text immediately after the `ending_sentence` of the previous scene. The final scene MUST have `ending_sentence` equal to the last sentence number of the chapter text.
2.  **Sentence Numbers:** Each sentence in the chapter text is labeled with a number in brackets (e.g., `[1]`, `[2]`, `[3]`). You must use **only the sentence number** (as an integer, without brackets) as the `ending_sentence` value. Do NOT use the full sentence text. For example, if a scene ends at the sentence labeled `[5]`, set `ending_sentence` to `5` (not the actual sentence text).

# 3. Output Instructions
Analyze the chapter text. Extract a list of scenes. For each scene, output a JSON object with the following fields:

* **`scene_title`**: A short, punchy, 3-6 word title capturing the essence of the scene.
* **`ending_sentence`**: The sentence number (as an integer) that marks the absolute end of this scene. This must be **only the number** from the label, not the full sentence text (e.g., if the sentence is labeled `[5]`, use `5`, not the actual sentence content).
* **`synopsis`**: A concise summary of the *external action* (Who did what? What events occurred?).
* **`significance`**: An analysis of the *internal importance*. Why does this scene exist? (e.g., reveals Character X's weakness, advances the mystery plot, establishes the theme of betrayal).

# 4. Final Output Format
Return ONLY a valid JSON array. Do not wrap the JSON in markdown code blocks or add conversational text.

Example output:
[
  {
    "scene_title": "The Argument in the Kitchen",
    "ending_sentence": 5,
    "synopsis": "John confronts Mary about the missing money. Mary denies it but reveals she lost her job.",
    "significance": "The scene shifts the dynamic from partnership to suspicion. It introduces the theme of financial insecurity and establishes Mary's tendency to lie under pressure."
  }
]

Now, analyze the following book chapter and extract the scenes.

<chapter>
{{{chapter_text}}}
</chapter>
\end{promptbox}
\captionof{table}{Prompt template for extracting scenes from a sentence-numbered chapter.}
\label{tab:prompt-scene-extraction}

\begin{promptbox}[title={Top-Down Storyline Prompt: First Level}]
You are an expert literary analyst specializing in narrative synthesis and thematic consolidation.

# Your Input
You will be provided with a chronological sequence of "Scenes" derived from a book. Each scene includes a `Synopsis` (what happens) and a `Significance` (why does this scene exist).

# Your Task
Analyze the provided scenes to identify the "Major Storylines". These are the pillars of the book--plots or character developments that the author dedicates the most content to. Such storylines should either involve major characters or convey a significant theme or idea that is important and recurrent throughout the book.

# Title & Description Requirements (CRITICAL)
- Titles must be straightforward and concrete:
  - Explicitly name the specific characters, factions, or groups involved.
  - Describe the specific situation or conflict using concrete actions/events.
  - Do NOT use abstract thematic labels or literary trope terms (e.g., "Coming of Age", "Redemption Arc", "Found Family").
  - Do NOT use poetic language, metaphors, analogies, or exaggeration.
  - Do NOT use a colon (:) in the title.

- Descriptions must be objective and specific:
  - Use formal, simple academic language.
  - Clearly state who is involved, what actions they take, key locations when relevant, and how the arc develops or changes over time.
  - Synthesize the plot events and, when useful, the narrative role indicated by the scene significance fields.
  - Do not introduce speculative literary interpretations beyond the provided scene information.
  - Avoid vague placeholders like "someone" or "the hero" unless the source text is ambiguous.

# Guidelines for Identification
A Major Storyline is a cohesive thread that runs through the book. It often falls into (but is not limited to) these categories:
1. Character Evolution: A specific person's growth, regression, or psychological change over time.
2. Interpersonal Dynamics: The trajectory of relationships (love, hate, mentorship, betrayal) between specific characters.
3. Central Conflict: Clashing values, Good vs. Evil, or political/military struggles between opposing parties.
4. Societal/Philosophical Discourse: How the narrative explores gender, religion, politics, or ethics through specific events.
5. World-Building (Speculative Fiction): The revelation of the imaginary world's mechanics and its intersection with reality or the characters.

# Output Structure
Output a single valid JSON object containing a list of storylines. Do not include markdown formatting or conversational text.

# Schema:
{
  "storylines": [
    {
      "title": "A concrete title containing specific character names and the specific nature of their interaction. Avoid using literary tropes or thematic labels.",
      "description": "A formal summary of the storyline that synthesizes relevant plot events and their narrative role from the scene synopsis and significance fields. Clearly state who is involved, what actions they take, and how the arc concludes or evolves."
    }
  ]
}

Now, analyze the following sequence of scenes and extract the major storylines.

<scenes>
{{{scenes}}}
</scenes>
\end{promptbox}
\captionof{table}{Prompt template for generating the first level of the top-down storyline tree.}
\label{tab:prompt-topdown-first}

\begin{promptbox}[title={Top-Down Storyline Prompt: Subsequent Levels}]
You are an expert literary analyst specializing in narrative synthesis and hierarchical storyline decomposition.

# Your Input
You will be given:
1) A chronological sequence of Scenes from a book. Each scene contains:
   - Synopsis: what happens
   - Significance: why the scene exists / what it contributes
2) A list of parent storylines that you must break into sub-storylines, each sub-storyline should include:
   - title
   - description

# Your Task
Produce child sub-storylines for the parent storylines in the input list that can be meaningfully decomposed. If a parent storyline is already minimal enough, omit it from the output rather than creating a decision field.

# Task Guidelines
- **Strict Hierarchy:** For each parent storyline, its sub-storylines must be directly related and contained within the parent storyline. Do not create sub-storylines that are unrelated to the context provided.
- **Granularity:** The sub-narratives must be narrower in scope than the parent storyline. For example, if the parent storyline is a war, a sub-narrative is a specific battle or a general's internal crisis.
- **Identify Common Sub-storylines:** If a sub-storyline can be considered as a common sub-storyline of multiple parent storylines, only output this sub-storyline once and link it to all the parent storylines that it is common to.
- **Size Constraints:** Produce 2-4 sub-storylines for each decomposed parent storyline. Do not decompose a parent storyline if it can be covered by less than 10 relevant scenes or if further splitting would point to individual scenes.

# Title & Description Requirements (CRITICAL)
- Titles must be straightforward and concrete:
  - Explicitly name the specific characters, factions, or groups involved.
  - Describe the specific situation or conflict using concrete actions/events.
  - Do NOT use abstract thematic labels or literary trope terms (e.g., "Coming of Age", "Redemption Arc", "Found Family").
  - Do NOT use poetic language, metaphors, analogies, or exaggeration.
  - Do NOT use a colon (:) in the title.

- Descriptions must be objective and specific:
  - Use formal, simple academic language.
  - Clearly state who is involved, what actions they take, key locations when relevant, and how the arc develops or changes over time.
  - Synthesize the plot events and, when useful, the narrative role indicated by the scene significance fields.
  - Do not introduce speculative literary interpretations beyond the provided scene information.
  - Avoid vague placeholders like "someone" or "the hero" unless the source text is ambiguous.

# Output Structure
Output a single valid JSON object containing a list of sub-storylines. Do not include markdown formatting or conversational text.

# Schema:
{
  "sub-storylines": [
    {
      "title": "A concrete title containing specific character names and the specific nature of their interaction. Avoid using literary tropes or thematic labels.",
      "description": "A formal summary that integrates relevant details from the parent storyline and the supporting scenes.",
      "parent_storylines": [<parent_storyline_number1>, <parent_storyline_number2>, ...]
    }
  ]
}

# parent_storylines (CRITICAL)
- The parent storylines above are numbered from 1 to n. In each sub-storyline, set "parent_storylines" to a list of **parent storyline numbers** (integers 1 to n) that this sub-storyline belongs under.
- Do NOT use the original parent storyline titles in "parent_storylines". Use only the corresponding numbers (e.g. [1], [2, 3], [1, 2, 4]).

Now, analyze the following input and produce the sub-storylines.

<parent_storylines>
{{{parent_storylines}}}
</parent_storylines>

<scenes>
{{{scenes}}}
</scenes>
\end{promptbox}
\captionof{table}{Prompt template for recursively generating lower levels of the top-down storyline tree.}
\label{tab:prompt-topdown-subsequent}

\begin{promptbox}[title={Adaptive Retrieval Prompt}]
You are a literature professor answering a question about a novel.

You are given the complete hierarchical storylines for the book (with Level, Storyline title, and Description). You may also be given an initial set of scenes retrieved directly from the question. Use any provided initial scenes as helpful supporting context, but treat them as incomplete and request retrieval from a storyline node whenever you still need better evidence.

You must always respond in exactly one of these two formats:

Reasoning: <thought about the problem>
Decision: retrieve
Storyline Title: <title of storyline node to retrieve>

or

Reasoning: <thought about the problem>
Decision: answer
Answer: \boxed{final answer}

Rules:
- Always output "Reasoning:" and "Decision:".
- If Decision is "retrieve", also output exactly one "Storyline Title:" field and do not output "Answer:".
- If Decision is "answer", also output exactly one "Answer:" field in the format `Answer: \boxed{<final answer>}`, and do not output "Storyline Title:".
- The `Storyline Title` should match a storyline title from the prompt after "Storyline:". Minor typos may be resolved via fuzzy matching.
- If initial scenes are provided, use them if they help, but do not assume they are sufficient.
- If the provided storylines are not sufficient, you must choose `Decision: retrieve` and provide the single best storyline title to retrieve from.
- Never answer with "not mentioned", "not found", or any equivalent abstention. If you do not have enough information yet, retrieve something.
- After each retrieval, the user will send a new message containing the retrieved scenes inside <scenes>...</scenes>. You may request more retrievals across turns until you can answer.
- Keep the final answer short (one or several words) unless the question clearly needs more.

Complete storylines for this book:

<storylines>
{{{storyline_text}}}
</storylines>

Initial scenes retrieved directly from the question (helpful but not guaranteed to be sufficient):

<initial_scenes>
{{{initial_scenes_text}}}
</initial_scenes>

Question: {{{question}}}

Use the storylines and any provided initial scenes to respond in the required structured format.
\end{promptbox}
\captionof{table}{Adaptive retrieval using storyline trees.}
\label{tab:prompt-adaptive-retrieval}

\end{document}

%% file: topdown2.tex

\definecolor{scenefill}{HTML}{F3DDD0}
\definecolor{sceneborder}{HTML}{B66A4A}

\definecolor{treefill}{HTML}{FAF4EB}
\definecolor{treeborder}{HTML}{B85A2A}
\definecolor{treeaccent}{HTML}{B85A2A}

\definecolor{edgegray}{HTML}{5F6770}
\definecolor{framegray}{HTML}{A5A099}
\definecolor{framefill}{HTML}{FCF8F2}
\definecolor{textgray}{HTML}{333333}

\resizebox{\linewidth}{!}{%
\begin{tikzpicture}[
    scale=.97,
    scenenode/.append style={
    minimum width=5mm,
    minimum height=6.8mm
    },
    font=\sffamily\small,
    treeedge/.style={
        draw=edgegray,
        line width=0.42pt
    },
    framebox/.style={
        draw=framegray,
        dashed,
        rounded corners=4pt,
        fill=framefill
    },
    scene/.style={
        draw=sceneborder,
        rounded corners=2pt,
        fill=scenefill,
        minimum width=1.16cm,
        minimum height=0.42cm,
        inner xsep=1pt,
        inner ysep=0.5pt,
        font=\sffamily\tiny
    },
    story/.style={
        circle,
        draw=treeborder,
        fill=treefill,
        minimum size=0.34cm,
        inner sep=0pt
    },
    lab/.style={
        font=\sffamily\scriptsize,
        align=center,
        text=black!85
    },
    smalllab/.style={
        font=\sffamily\scriptsize,
        align=left,
        text=edgegray
    },
    sample/.style={
        draw=treeborder!75,
        rounded corners=3pt,
        fill=white,
        text width=5.55cm,
        minimum height=0.84cm,
        inner sep=3.6pt,
        align=left,
        font=\sffamily\fontsize{5.0pt}{5.6pt}\selectfont
    },
    expandarrow/.style={
        -{Latex[length=1.7mm]},
        draw=edgegray,
        line width=0.50pt
    },
    curvearrow/.style={
        -{Latex[length=1.5mm]},
        dashed,
        draw=edgegray,
        line width=0.40pt
    }
]

\node[
    lab,
    font=\sffamily\scriptsize\bfseries,
    text=black!85
] (title) at (4.95,2.55)
{Top-down storyline tree construction};

\node[scene] (s1) at (2.90,1.94) {Scene $1$};
\node[scene] (s2) at (4.25,1.94) {Scene $2$};
\node[smalllab, align=center] (sdots) at (5.60,1.94) {$\cdots$};
\node[scene] (sN) at (6.95,1.94) {Scene $N$};

\node[smalllab, anchor=west] (labS) at (8.10,1.94)
{$\mathcal{S}$};

\coordinate (expandTop) at (1.62,1.18);
\coordinate (expandBot) at (1.62,-1.12);

\draw[expandarrow] (expandTop) -- (expandBot);

\node[
    font=\sffamily\tiny,
    rotate=90,
    text=edgegray
] (expandLabel) at (1.49,0.03) {expand};

\coordinate (root) at (4.95,1.48);

\node[story] (t11) at (3.80,0.65) {};
\node[story] (t12) at (4.95,0.65) {};
\node[story] (t13) at (6.10,0.65) {};

\node[story] (t21) at (3.20,-0.25) {};
\node[story] (t22) at (4.05,-0.25) {};
\node[story] (t23) at (4.65,-0.25) {};
\node[story] (t24) at (5.25,-0.25) {};
\node[story] (t25) at (5.85,-0.25) {};
\node[story] (t26) at (6.70,-0.25) {};

\node[story] (t31) at (2.70,-1.20) {};
\node[story] (t32) at (3.26,-1.20) {};
\node[story] (t33) at (3.83,-1.20) {};
\node[story] (t34) at (4.39,-1.20) {};
\node[story] (t35) at (4.95,-1.20) {};
\node[story] (t36) at (5.51,-1.20) {};
\node[story] (t37) at (6.08,-1.20) {};
\node[story] (t38) at (6.64,-1.20) {};
\node[story] (t39) at (7.20,-1.20) {};

\draw[treeedge] (root) -- (t11);
\draw[treeedge] (root) -- (t12);
\draw[treeedge] (root) -- (t13);

\draw[treeedge] (t11) -- (t21);
\draw[treeedge] (t11) -- (t22);

\draw[treeedge] (t12) -- (t23);
\draw[treeedge] (t12) -- (t24);

\draw[treeedge] (t13) -- (t25);
\draw[treeedge] (t13) -- (t26);

\draw[treeedge] (t21) -- (t31);
\draw[treeedge] (t21) -- (t32);

\draw[treeedge] (t22) -- (t33);

\draw[treeedge] (t23) -- (t34);
\draw[treeedge] (t23) -- (t35);

\draw[treeedge] (t24) -- (t36);

\draw[treeedge] (t25) -- (t37);

\draw[treeedge] (t26) -- (t38);
\draw[treeedge] (t26) -- (t39);

\node[smalllab, anchor=west] (l1) at (8.10,0.65)
{$\mathcal{T}^{(1)}$};

\node[smalllab, anchor=west] (l2) at (8.10,-0.25)
{$\mathcal{T}^{(2)}$};

\node[smalllab, anchor=west] (l3) at (8.10,-1.20)
{$\mathcal{T}^{(3)}$};

\begin{pgfonlayer}{background}
\node[
    framebox,
    inner xsep=5pt,
    inner ysep=2pt,
    fit=(title)(s1)(sN)(labS)(expandTop)(expandBot)(expandLabel)(t31)(t39)(l1)(l3)
] (processbox) {};
\end{pgfonlayer}

\node[sample] (samplebox) at (4.95,-2.52) {
{\sffamily\fontsize{5.9pt}{6.3pt}\selectfont\bfseries\textcolor{treeaccent}{Sample storyline node}}\\[1.1pt]
\textbf{Title:} Heathcliff Returns Changed and Wealthy\\
\textbf{Description:} After a long absence, Heathcliff suddenly reappears \ldots\ He is no longer the powerless boy of the house, but a man carrying pride and revenge.
};

\draw[curvearrow]
(samplebox.north) to[out=70,in=-105] (t36.south);

\end{tikzpicture}%
}

%% file: bottomup2.tex

\definecolor{scenefill}{HTML}{F3DDD0}
\definecolor{sceneborder}{HTML}{B66A4A}

\definecolor{treefill}{HTML}{FAF4EB}
\definecolor{treeborder}{HTML}{B85A2A}
\definecolor{treeaccent}{HTML}{B85A2A}

\definecolor{edgegray}{HTML}{5F6770}
\definecolor{framegray}{HTML}{A5A099}
\definecolor{framefill}{HTML}{FCF8F2}
\definecolor{textgray}{HTML}{333333}

\resizebox{\linewidth}{!}{%
\begin{tikzpicture}[
    font=\sffamily\small,
    treeedge/.style={
        line width=0.42pt,
        draw=edgegray
    },
    sceneedge/.style={
        line width=0.38pt,
        draw=edgegray,
        dashed
    },
    processarrow/.style={
        -{Latex[length=1.55mm]},
        line width=0.42pt,
        draw=edgegray
    },
    scene/.style={
        draw=sceneborder,
        rounded corners=1.5pt,
        fill=scenefill,
        minimum width=0.44cm,
        minimum height=0.24cm,
        inner sep=0pt
    },
    story/.style={
        circle,
        draw=treeborder,
        fill=treefill,
        minimum size=0.34cm,
        inner sep=0pt
    },
    root/.style={
        circle,
        draw=treeborder,
        fill=treefill,
        minimum size=0.37cm,
        inner sep=0pt
    },
    lab/.style={
        font=\sffamily\scriptsize,
        align=center,
        text=black!85
    },
    smalllab/.style={
        font=\sffamily\scriptsize,
        align=left,
        text=edgegray
    },
    framebox/.style={
        draw=framegray,
        dashed,
        rounded corners=3pt,
        fill=framefill
    }
]

\node[
    lab,
    font=\sffamily\scriptsize\bfseries,
    text=black!85
] (title) at (4.05,2.03)
{Bottom-up storyline tree construction};

\draw[processarrow] (0.90,-0.66) -- (0.90,1.35);

\node[
    font=\sffamily\tiny,
    rotate=90,
    text=edgegray
] (processlabel) at (0.72,0.35) {cluster \& summarize};

\node[root] (t11) at (3.05,1.35) {};
\node[root] (t12) at (4.05,1.35) {};
\node[root] (t13) at (5.05,1.35) {};

\node[story] (t21) at (2.35,0.60) {};
\node[story] (t22) at (3.05,0.60) {};
\node[story] (t23) at (3.76,0.60) {};
\node[story] (t24) at (4.48,0.60) {};
\node[story] (t25) at (5.05,0.60) {};
\node[story] (t26) at (5.66,0.60) {};

\node[story] (t31) at (2.100,-0.15) {};
\node[story] (t32) at (2.575,-0.15) {};
\node[story] (t33) at (3.050,-0.15) {};
\node[story] (t34) at (3.525,-0.15) {};
\node[story] (t35) at (4.000,-0.15) {};
\node[story] (t36) at (4.475,-0.15) {};
\node[story] (t37) at (4.950,-0.15) {};
\node[story] (t38) at (5.425,-0.15) {};
\node[story] (t39) at (5.900,-0.15) {};

\node[scene] (s1) at (1.85,-0.95) {};
\node[scene] (s2) at (2.40,-0.95) {};
\node[scene] (s3) at (2.95,-0.95) {};
\node[scene] (s4) at (3.50,-0.95) {};
\node[smalllab, align=center] (sdots) at (4.05,-0.95) {$\cdots$};
\node[scene] (s5) at (4.60,-0.95) {};
\node[scene] (s6) at (5.15,-0.95) {};
\node[scene] (s7) at (5.70,-0.95) {};
\node[scene] (s8) at (6.25,-0.95) {};

\draw[treeedge] (t11) -- (t21);
\draw[treeedge] (t11) -- (t22);
\draw[treeedge] (t12) -- (t23);
\draw[treeedge] (t12) -- (t24);
\draw[treeedge] (t13) -- (t25);
\draw[treeedge] (t13) -- (t26);

\draw[treeedge] (t21) -- (t31);
\draw[treeedge] (t21) -- (t32);
\draw[treeedge] (t22) -- (t33);
\draw[treeedge] (t23) -- (t34);
\draw[treeedge] (t23) -- (t35);
\draw[treeedge] (t24) -- (t36);
\draw[treeedge] (t25) -- (t37);
\draw[treeedge] (t26) -- (t38);
\draw[treeedge] (t26) -- (t39);

\draw[sceneedge] (t31) -- (s1);
\draw[sceneedge] (t32) -- (s2);
\draw[sceneedge] (t33) -- (s3);
\draw[sceneedge] (t34) -- (s4);

\coordinate (sdotsTop) at (4.05,-0.83);
\draw[sceneedge] (t35) -- (sdotsTop);

\draw[sceneedge] (t36) -- (s5);
\draw[sceneedge] (t37) -- (s6);
\draw[sceneedge] (t38) -- (s7);
\draw[sceneedge] (t39) -- (s8);

\node[smalllab, anchor=west] (l1) at (6.90,1.35)
{$\mathcal{T}^{(1)}$};

\node[smalllab, anchor=west] (l2) at (6.90,0.60)
{$\mathcal{T}^{(2)}$};

\node[smalllab, anchor=west] (l3) at (6.90,-0.15)
{$\mathcal{T}^{(3)}$};

\node[smalllab, anchor=west] (l4) at (6.90,-0.95)
{$\mathcal{T}^{(L+1)}=\mathcal{S}$};

\begin{pgfonlayer}{background}
\node[
    framebox,
    inner xsep=5pt,
    inner ysep=4pt,
    fit=(title)(processlabel)(t11)(t13)(s1)(s8)(l1)(l4)
] (processbox) {};
\end{pgfonlayer}

\end{tikzpicture}%
}

%% file: custom.bib
@misc{qu2024semanticchunking,
      title={Is Semantic Chunking Worth the Computational Cost?}, 
      author={Renyi Qu and Ruixuan Tu and Forrest Bao},
      year={2024},
      eprint={2410.13070},
      archivePrefix={arXiv},
      primaryClass={cs.CL},
      url={https://arxiv.org/abs/2410.13070}, 
}

@misc{chandak2025answermatching,
      title={Answer Matching Outperforms Multiple Choice for Language Model Evaluation}, 
      author={Nikhil Chandak and Shashwat Goel and Ameya Prabhu and Moritz Hardt and Jonas Geiping},
      year={2025},
      eprint={2507.02856},
      archivePrefix={arXiv},
      primaryClass={cs.CL},
      url={https://arxiv.org/abs/2507.02856}, 
}

@misc{xu2025detectiveqa,
      title={DetectiveQA: Evaluating Long-Context Reasoning on Detective Novels}, 
      author={Zhe Xu and Jiasheng Ye and Xiaoran Liu and Xiangyang Liu and Tianxiang Sun and Zhigeng Liu and Qipeng Guo and Linlin Li and Qun Liu and Xuanjing Huang and Xipeng Qiu},
      year={2025},
      eprint={2409.02465},
      archivePrefix={arXiv},
      primaryClass={cs.CL},
      url={https://arxiv.org/abs/2409.02465}, 
}

@article{santana2023survey,
  title = {A Survey on Narrative Extraction from Textual Data},
  author = {Santana, Brenda and Campos, Ricardo and Amorim, Evelin and Jorge, Al{\'i}pio and Silvano, Purifica{\c c}{\~a}o and Nunes, S{\'e}rgio},
  journal = {Artificial Intelligence Review},
  volume = {56},
  pages = {8393--8435},
  year = {2023},
  doi = {10.1007/s10462-022-10338-7},
  url = {https://link.springer.com/article/10.1007/s10462-022-10338-7}
}

@inproceedings{papalampidi2019movieplot,
    title = "Movie Plot Analysis via Turning Point Identification",
    author = "Papalampidi, Pinelopi  and
      Keller, Frank  and
      Lapata, Mirella",
    editor = "Inui, Kentaro  and
      Jiang, Jing  and
      Ng, Vincent  and
      Wan, Xiaojun",
    booktitle = "Proceedings of the 2019 Conference on Empirical Methods in Natural Language Processing and the 9th International Joint Conference on Natural Language Processing (EMNLP-IJCNLP)",
    month = nov,
    year = "2019",
    address = "Hong Kong, China",
    publisher = "Association for Computational Linguistics",
    url = "https://aclanthology.org/D19-1180/",
    doi = "10.18653/v1/D19-1180",
    pages = "1707--1717"
}

@inproceedings{jin2025hierarchicalrefinement,
    title = "Hierarchical Document Refinement for Long-context Retrieval-augmented Generation",
    author = "Jin, Jiajie  and
      Li, Xiaoxi  and
      Dong, Guanting  and
      Zhang, Yuyao  and
      Zhu, Yutao  and
      Wu, Yongkang  and
      Li, Zhonghua  and
      Qi, Ye  and
      Dou, Zhicheng",
    editor = "Che, Wanxiang  and
      Nabende, Joyce  and
      Shutova, Ekaterina  and
      Pilehvar, Mohammad Taher",
    booktitle = "Proceedings of the 63rd Annual Meeting of the Association for Computational Linguistics (Volume 1: Long Papers)",
    month = jul,
    year = "2025",
    address = "Vienna, Austria",
    publisher = "Association for Computational Linguistics",
    url = "https://aclanthology.org/2025.acl-long.176/",
    doi = "10.18653/v1/2025.acl-long.176",
    pages = "3502--3520",
    ISBN = "979-8-89176-251-0"
}

@inproceedings{tao2025treerag,
    title = "{T}ree{RAG}: Unleashing the Power of Hierarchical Storage for Enhanced Knowledge Retrieval in Long Documents",
    author = "Tao, Wenyu  and
      Xing, Xiaofen  and
      Chen, Yirong  and
      Huang, Linyi  and
      Xu, Xiangmin",
    editor = "Che, Wanxiang  and
      Nabende, Joyce  and
      Shutova, Ekaterina  and
      Pilehvar, Mohammad Taher",
    booktitle = "Findings of the Association for Computational Linguistics: ACL 2025",
    month = jul,
    year = "2025",
    address = "Vienna, Austria",
    publisher = "Association for Computational Linguistics",
    url = "https://aclanthology.org/2025.findings-acl.20/",
    doi = "10.18653/v1/2025.findings-acl.20",
    pages = "356--371",
    ISBN = "979-8-89176-256-5"
}

@inproceedings{zehe2021detectingscenes,
    title = "Detecting Scenes in Fiction: A new Segmentation Task",
    author = {Zehe, Albin  and
      Konle, Leonard  and
      D{\"u}mpelmann, Lea Katharina  and
      Gius, Evelyn  and
      Hotho, Andreas  and
      Jannidis, Fotis  and
      Kaufmann, Lucas  and
      Krug, Markus  and
      Puppe, Frank  and
      Reiter, Nils  and
      Schreiber, Annekea  and
      Wiedmer, Nathalie},
    editor = "Merlo, Paola  and
      Tiedemann, Jorg  and
      Tsarfaty, Reut",
    booktitle = "Proceedings of the 16th Conference of the European Chapter of the Association for Computational Linguistics: Main Volume",
    month = apr,
    year = "2021",
    address = "Online",
    publisher = "Association for Computational Linguistics",
    url = "https://aclanthology.org/2021.eacl-main.276/",
    doi = "10.18653/v1/2021.eacl-main.276",
    pages = "3167--3177"
}

@misc{zhang2024coa,
      title={Chain of Agents: Large Language Models Collaborating on Long-Context Tasks}, 
      author={Yusen Zhang and Ruoxi Sun and Yanfei Chen and Tomas Pfister and Rui Zhang and Sercan {\"O}. Arik},
      year={2024},
      eprint={2406.02818},
      archivePrefix={arXiv},
      primaryClass={cs.CL},
      url={https://arxiv.org/abs/2406.02818}, 
}

@inproceedings{gaizauskas2015sceneml,
  title = {{SceneML}: A Proposal for Annotating Scenes in Narrative Text},
  author = {Gaizauskas, Robert and Alrashid, Tarfah},
  editor = {Bunt, Harry},
  booktitle = {Proceedings of the Fifteenth Joint ACL - ISO Workshop on Interoperable Semantic Annotation (ISA-15)},
  month = may,
  year = {2019},
  address = {Gothenburg, Sweden},
  publisher = {TiCC, Tilburg center for Cognition and Communication},
  pages = {13--21},
  url = {https://sigsem.uvt.nl/isa15/ISA-15_proceedings.pdf}
}

@article{liu2023lostinmiddle,
    title = "Lost in the Middle: How Language Models Use Long Contexts",
    author = "Liu, Nelson F.  and
      Lin, Kevin  and
      Hewitt, John  and
      Paranjape, Ashwin  and
      Bevilacqua, Michele  and
      Petroni, Fabio  and
      Liang, Percy",
    journal = "Transactions of the Association for Computational Linguistics",
    volume = "12",
    year = "2024",
    address = "Cambridge, MA",
    publisher = "MIT Press",
    url = "https://aclanthology.org/2024.tacl-1.9/",
    doi = "10.1162/tacl_a_00638",
    pages = "157--173"
}

@misc{sarthi2024raptor,
  title = {{RAPTOR}: Recursive Abstractive Processing for Tree-Organized Retrieval},
  author = {Parth Sarthi and Salman Abdullah and Aditi Tuli and Shubh Khanna and Anna Goldie and Christopher D. Manning},
  year = {2024},
  eprint = {2401.18059},
  archivePrefix = {arXiv},
  primaryClass = {cs.CL},
  doi = {10.48550/arXiv.2401.18059},
  url = {https://arxiv.org/abs/2401.18059}
}

@misc{zheng2023llmjudge,
      title={Judging LLM-as-a-Judge with MT-Bench and Chatbot Arena}, 
      author={Lianmin Zheng and Wei-Lin Chiang and Ying Sheng and Siyuan Zhuang and Zhanghao Wu and Yonghao Zhuang and Zi Lin and Zhuohan Li and Dacheng Li and Eric P. Xing and Hao Zhang and Joseph E. Gonzalez and Ion Stoica},
      year={2023},
      eprint={2306.05685},
      archivePrefix={arXiv},
      primaryClass={cs.CL},
      url={https://arxiv.org/abs/2306.05685}, 
}

@misc{zhang2025qwen3embedding,
      title={Qwen3 Embedding: Advancing Text Embedding and Reranking Through Foundation Models}, 
      author={Yanzhao Zhang and Mingxin Li and Dingkun Long and Xin Zhang and Huan Lin and Baosong Yang and Pengjun Xie and An Yang and Dayiheng Liu and Junyang Lin and Fei Huang and Jingren Zhou},
      year={2025},
      eprint={2506.05176},
      archivePrefix={arXiv},
      primaryClass={cs.CL},
      url={https://arxiv.org/abs/2506.05176}, 
}

@misc{yang2025qwen3,
      title={Qwen3 Technical Report}, 
      author={An Yang and Anfeng Li and Baosong Yang and Beichen Zhang and Binyuan Hui and Bo Zheng and Bowen Yu and Chang Gao and Chengen Huang and Chenxu Lv and Chujie Zheng and Dayiheng Liu and Fan Zhou and Fei Huang and Feng Hu and Hao Ge and Haoran Wei and Huan Lin and Jialong Tang and Jian Yang and Jianhong Tu and Jianwei Zhang and Jianxin Yang and Jiaxi Yang and Jing Zhou and Jingren Zhou and Junyang Lin and Kai Dang and Keqin Bao and Kexin Yang and Le Yu and Lianghao Deng and Mei Li and Mingfeng Xue and Mingze Li and Pei Zhang and Peng Wang and Qin Zhu and Rui Men and Ruize Gao and Shixuan Liu and Shuang Luo and Tianhao Li and Tianyi Tang and Wenbiao Yin and Xingzhang Ren and Xinyu Wang and Xinyu Zhang and Xuancheng Ren and Yang Fan and Yang Su and Yichang Zhang and Yinger Zhang and Yu Wan and Yuqiong Liu and Zekun Wang and Zeyu Cui and Zhenru Zhang and Zhipeng Zhou and Zihan Qiu},
      year={2025},
      eprint={2505.09388},
      archivePrefix={arXiv},
      primaryClass={cs.CL},
      url={https://arxiv.org/abs/2505.09388}, 
}

@article{zwaan1995eventindexing,
 ISSN = {09567976, 14679280},
 URL = {http://www.jstor.org/stable/40063035},
 author = {Rolf A. Zwaan and Mark C. Langston and Arthur C. Graesser},
 journal = {Psychological Science},
 number = {5},
 pages = {292--297},
 publisher = {[Association for Psychological Science, Sage Publications, Inc.]},
 title = {The Construction of Situation Models in Narrative Comprehension: An Event-Indexing Model},
 urldate = {2026-05-23},
 volume = {6},
 year = {1995}
}

@misc{yu2025memagent,
      title={MemAgent: Reshaping Long-Context LLM with Multi-Conv RL-based Memory Agent}, 
      author={Hongli Yu and Tinghong Chen and Jiangtao Feng and Jiangjie Chen and Weinan Dai and Qiying Yu and Ya-Qin Zhang and Wei-Ying Ma and Jingjing Liu and Mingxuan Wang and Hao Zhou},
      year={2025},
      eprint={2507.02259},
      archivePrefix={arXiv},
      primaryClass={cs.CL},
      url={https://arxiv.org/abs/2507.02259}, 
}

@misc{yu2025treeagents,
      title={Tree of Agents: Improving Long-Context Capabilities of Large Language Models through Multi-Perspective Reasoning}, 
      author={Song Yu and Xiaofei Xu and Ke Deng and Li Li and Lin Tian},
      year={2025},
      eprint={2509.06436},
      archivePrefix={arXiv},
      primaryClass={cs.AI},
      url={https://arxiv.org/abs/2509.06436}, 
}

@article{zwaan1998situationmodels,
  title = {Situation Models in Language Comprehension and Memory},
  author = {Zwaan, Rolf A. and Radvansky, Gabriel A.},
  journal = {Psychological Bulletin},
  volume = {123},
  number = {2},
  pages = {162--185},
  year = {1998},
  doi = {10.1037/0033-2909.123.2.162},
  url = {https://psycnet.apa.org/doi/10.1037/0033-2909.123.2.162}
}

@article{zacks2007eventperception,
  title = {Event Perception: A Mind-Brain Perspective},
  author = {Zacks, Jeffrey M. and Speer, Nicole K. and Swallow, Khena M. and Braver, Todd S. and Reynolds, Jeremy R.},
  journal = {Psychological Bulletin},
  volume = {133},
  number = {2},
  pages = {273--293},
  year = {2007},
  doi = {10.1037/0033-2909.133.2.273},
  url = {https://doi.org/10.1037/0033-2909.133.2.273}
}

@misc{hsieh2024ruler,
  title = {{RULER}: What's the Real Context Size of Your Long-Context Language Models?},
  author = {Hsieh, Cheng-Ping and Sun, Simeng and Kriman, Samuel and Acharya, Shantanu and Rekesh, Dima and Jia, Fei and Zhang, Yang and Ginsburg, Boris},
  year = {2024},
  eprint = {2404.06654},
  archivePrefix = {arXiv},
  primaryClass = {cs.CL},
  doi = {10.48550/arXiv.2404.06654},
  url = {https://arxiv.org/abs/2404.06654}
}

@misc{modarressi2025nolima,
      title={NoLiMa: Long-Context Evaluation Beyond Literal Matching}, 
      author={Ali Modarressi and Hanieh Deilamsalehy and Franck Dernoncourt and Trung Bui and Ryan A. Rossi and Seunghyun Yoon and Hinrich Schütze},
      year={2025},
      eprint={2502.05167},
      archivePrefix={arXiv},
      primaryClass={cs.CL},
      url={https://arxiv.org/abs/2502.05167}, 
}

@inproceedings{glavas2016semanticsegmentation,
    title = "Unsupervised Text Segmentation Using Semantic Relatedness Graphs",
    author = "Glava{\v{s}}, Goran  and
      Nanni, Federico  and
      Ponzetto, Simone Paolo",
    editor = "Gardent, Claire  and
      Bernardi, Raffaella  and
      Titov, Ivan",
    booktitle = "Proceedings of the Fifth Joint Conference on Lexical and Computational Semantics",
    month = aug,
    year = "2016",
    address = "Berlin, Germany",
    publisher = "Association for Computational Linguistics",
    url = "https://aclanthology.org/S16-2016/",
    doi = "10.18653/v1/S16-2016",
    pages = "125--130"
}

@article{kocisky2018narrativeqa,
  title = {The {NarrativeQA} Reading Comprehension Challenge},
  author = {Ko{\v{c}}isk{\'y}, Tom{\'a}{\v{s}} and Schwarz, Jonathan and Blunsom, Phil and Dyer, Chris and Hermann, Karl Moritz and Melis, G{\'a}bor and Grefenstette, Edward},
  journal = {Transactions of the Association for Computational Linguistics},
  volume = {6},
  year = {2018},
  address = {Cambridge, MA},
  publisher = {MIT Press},
  url = {https://aclanthology.org/Q18-1023/},
  doi = {10.1162/tacl_a_00023},
  pages = {317--328}
}

@misc{xu2025divideconquer,
      title={When Does Divide and Conquer Work for Long Context LLM? A Noise Decomposition Framework}, 
      author={Zhen Xu and Shang Zhu and Jue Wang and Junlin Wang and Ben Athiwaratkun and Chi Wang and James Zou and Ce Zhang},
      year={2025},
      eprint={2506.16411},
      archivePrefix={arXiv},
      primaryClass={cs.CL},
      url={https://arxiv.org/abs/2506.16411}, 
}

@inproceedings{tian2025distance,
    title = "Distance between Relevant Information Pieces Causes Bias in Long-Context {LLM}s",
    author = "Tian, Runchu  and
      Li, Yanghao  and
      Fu, Yuepeng  and
      Deng, Siyang  and
      Luo, Qinyu  and
      Qian, Cheng  and
      Wang, Shuo  and
      Cong, Xin  and
      Zhang, Zhong  and
      Wu, Yesai  and
      Lin, Yankai  and
      Wang, Huadong  and
      Liu, Xiaojiang",
    editor = "Che, Wanxiang  and
      Nabende, Joyce  and
      Shutova, Ekaterina  and
      Pilehvar, Mohammad Taher",
    booktitle = "Findings of the Association for Computational Linguistics: ACL 2025",
    month = jul,
    year = "2025",
    address = "Vienna, Austria",
    publisher = "Association for Computational Linguistics",
    url = "https://aclanthology.org/2025.findings-acl.28/",
    doi = "10.18653/v1/2025.findings-acl.28",
    pages = "521--533",
    ISBN = "979-8-89176-256-5"
}

@misc{veseli2025positional,
      title={Positional Biases Shift as Inputs Approach Context Window Limits}, 
      author={Blerta Veseli and Julian Chibane and Mariya Toneva and Alexander Koller},
      year={2025},
      eprint={2508.07479},
      archivePrefix={arXiv},
      primaryClass={cs.CL},
      url={https://arxiv.org/abs/2508.07479}, 
}

@article{schwarz1978estimating,
 ISSN = {00905364, 21688966},
 URL = {http://www.jstor.org/stable/2958889},
 author = {Gideon Schwarz},
 journal = {The Annals of Statistics},
 number = {2},
 pages = {461--464},
 publisher = {Institute of Mathematical Statistics},
 title = {Estimating the Dimension of a Model},
 urldate = {2026-05-15},
 volume = {6},
 year = {1978}
}

@inproceedings{du2025context,
    title = "Context Length Alone Hurts {LLM} Performance Despite Perfect Retrieval",
    author = "Du, Yufeng  and
      Tian, Minyang  and
      Ronanki, Srikanth  and
      Rongali, Subendhu  and
      Bodapati, Sravan Babu  and
      Galstyan, Aram  and
      Wells, Azton  and
      Schwartz, Roy  and
      Huerta, Eliu A  and
      Peng, Hao",
    editor = "Christodoulopoulos, Christos  and
      Chakraborty, Tanmoy  and
      Rose, Carolyn  and
      Peng, Violet",
    booktitle = "Findings of the Association for Computational Linguistics: EMNLP 2025",
    month = nov,
    year = "2025",
    address = "Suzhou, China",
    publisher = "Association for Computational Linguistics",
    url = "https://aclanthology.org/2025.findings-emnlp.1264/",
    doi = "10.18653/v1/2025.findings-emnlp.1264",
    pages = "23281--23298",
    ISBN = "979-8-89176-335-7"
}

@inproceedings{kryscinski2022booksum,
  title = {{BOOKSUM}: A Collection of Datasets for Long-form Narrative Summarization},
  author = {Kryscinski, Wojciech and Rajani, Nazneen and Agarwal, Divyansh and Xiong, Caiming and Radev, Dragomir},
  booktitle = {Findings of the Association for Computational Linguistics: EMNLP 2022},
  month = dec,
  year = {2022},
  address = {Abu Dhabi, United Arab Emirates},
  publisher = {Association for Computational Linguistics},
  pages = {6536--6558},
  doi = {10.18653/v1/2022.findings-emnlp.488},
  url = {https://aclanthology.org/2022.findings-emnlp.488/}
}

@misc{shen2025qwenlongl15,
      title={QwenLong-L1.5: Post-Training Recipe for Long-Context Reasoning and Memory Management}, 
      author={Weizhou Shen and Ziyi Yang and Chenliang Li and Zhiyuan Lu and Miao Peng and Huashan Sun and Yingcheng Shi and Shengyi Liao and Shaopeng Lai and Bo Zhang and Dayiheng Liu and Fei Huang and Jingren Zhou and Ming Yan},
      year={2025},
      eprint={2512.12967},
      archivePrefix={arXiv},
      primaryClass={cs.CL},
      url={https://arxiv.org/abs/2512.12967}, 
}

@misc{zhang2026docdancer,
      title={DocDancer: Towards Agentic Document-Grounded Information Seeking}, 
      author={Qintong Zhang and Xinjie Lv and Jialong Wu and Baixuan Li and Zhengwei Tao and Guochen Yan and Huanyao Zhang and Bin Wang and Jiahao Xu and Haitao Mi and Wentao Zhang},
      year={2026},
      eprint={2601.05163},
      archivePrefix={arXiv},
      primaryClass={cs.CL},
      url={https://arxiv.org/abs/2601.05163}, 
}

@misc{wang2025novelqa,
      title={NovelQA: Benchmarking Question Answering on Documents Exceeding 200K Tokens}, 
      author={Cunxiang Wang and Ruoxi Ning and Boqi Pan and Tonghui Wu and Qipeng Guo and Cheng Deng and Guangsheng Bao and Xiangkun Hu and Zheng Zhang and Qian Wang and Yue Zhang},
      year={2025},
      eprint={2403.12766},
      archivePrefix={arXiv},
      primaryClass={cs.CL},
      url={https://arxiv.org/abs/2403.12766}, 
}

@inproceedings{bonomo2025literaryqa,
  title = {{LiteraryQA}: Towards Effective Evaluation of Long-document Narrative {QA}},
  author = {Bonomo, Tommaso and Gioffr{\'e}, Luca and Navigli, Roberto},
  booktitle = {Proceedings of the 2025 Conference on Empirical Methods in Natural Language Processing},
  month = nov,
  year = {2025},
  address = {Suzhou, China},
  publisher = {Association for Computational Linguistics},
  pages = {34086--34107},
  doi = {10.18653/v1/2025.emnlp-main.1729},
  url = {https://aclanthology.org/2025.emnlp-main.1729/}
}

@misc{zhou2025contextedus,
  title = {From Context to {EDU}s: Faithful and Structured Context Compression via Elementary Discourse Unit Decomposition},
  author = {Zhou, Yiqing and Lei, Yu and Si, Shuzheng and Sun, Qingyan and Wang, Wei and Wu, Yifei and Wen, Hao and Chen, Gang and Qi, Fanchao and Sun, Maosong},
  year = {2025},
  eprint = {2512.14244},
  archivePrefix = {arXiv},
  primaryClass = {cs.CL},
  doi = {10.48550/arXiv.2512.14244},
  url = {https://arxiv.org/abs/2512.14244}
}

@article{liu2026semanticraptor,
  author = {Liu, Yan and Xie, Xiaodong and Wan, Xin and Pan, Yi and Wang, Cheng},
  title = {Enhancing RAPTOR with Semantic Chunking and Adaptive Graph Clustering},
  journal = {Frontiers in Computer Science},
  volume = {7},
  eid = {1710121},
  year = {2026},
  url = {https://www.frontiersin.org/journals/computer-science/articles/10.3389/fcomp.2025.1710121},
  doi = {10.3389/fcomp.2025.1710121},
  issn = {2624-9898}
}

@misc{badshah2025referenceguidedverdict,
      title={Reference-Guided Verdict: LLMs-as-Judges in Automatic Evaluation of Free-Form QA}, 
      author={Sher Badshah and Hassan Sajjad},
      year={2025},
      eprint={2408.09235},
      archivePrefix={arXiv},
      primaryClass={cs.CL},
      url={https://arxiv.org/abs/2408.09235}, 
}

@misc{ho2025extractiveqajudge,
      title={LLM-as-a-Judge: Reassessing the Performance of LLMs in Extractive QA}, 
      author={Xanh Ho and Jiahao Huang and Florian Boudin and Akiko Aizawa},
      year={2025},
      eprint={2504.11972},
      archivePrefix={arXiv},
      primaryClass={cs.CL},
      url={https://arxiv.org/abs/2504.11972}, 
}

@article{guo2025deepseekr1,
  title={DeepSeek-R1 incentivizes reasoning in LLMs through reinforcement learning},
  volume={645},
  ISSN={1476-4687},
  url={http://dx.doi.org/10.1038/s41586-025-09422-z},
  DOI={10.1038/s41586-025-09422-z},
  number={8081},
  journal={Nature},
  publisher={Springer Science and Business Media LLC},
  author={Guo, Daya and Yang, Dejian and Zhang, Haowei and Song, Junxiao and Wang, Peiyi and Zhu, Qihao and Xu, Runxin and Zhang, Ruoyu and Ma, Shirong and Bi, Xiao and Zhang, Xiaokang and Yu, Xingkai and Wu, Yu and Wu, Z. F. and Gou, Zhibin and Shao, Zhihong and Li, Zhuoshu and Gao, Ziyi and Liu, Aixin and Xue, Bing and Wang, Bingxuan and Wu, Bochao and Feng, Bei and Lu, Chengda and Zhao, Chenggang and Deng, Chengqi and Ruan, Chong and Dai, Damai and Chen, Deli and Ji, Dongjie and Li, Erhang and Lin, Fangyun and Dai, Fucong and Luo, Fuli and Hao, Guangbo and Chen, Guanting and Li, Guowei and Zhang, H. and Xu, Hanwei and Ding, Honghui and Gao, Huazuo and Qu, Hui and Li, Hui and Guo, Jianzhong and Li, Jiashi and Chen, Jingchang and Yuan, Jingyang and Tu, Jinhao and Qiu, Junjie and Li, Junlong and Cai, J. L. and Ni, Jiaqi and Liang, Jian and Chen, Jin and Dong, Kai and Hu, Kai and You, Kaichao and Gao, Kaige and Guan, Kang and Huang, Kexin and Yu, Kuai and Wang, Lean and Zhang, Lecong and Zhao, Liang and Wang, Litong and Zhang, Liyue and Xu, Lei and Xia, Leyi and Zhang, Mingchuan and Zhang, Minghua and Tang, Minghui and Zhou, Mingxu and Li, Meng and Wang, Miaojun and Li, Mingming and Tian, Ning and Huang, Panpan and Zhang, Peng and Wang, Qiancheng and Chen, Qinyu and Du, Qiushi and Ge, Ruiqi and Zhang, Ruisong and Pan, Ruizhe and Wang, Runji and Chen, R. J. and Jin, R. L. and Chen, Ruyi and Lu, Shanghao and Zhou, Shangyan and Chen, Shanhuang and Ye, Shengfeng and Wang, Shiyu and Yu, Shuiping and Zhou, Shunfeng and Pan, Shuting and Li, S. S. and Zhou, Shuang and Wu, Shaoqing and Yun, Tao and Pei, Tian and Sun, Tianyu and Wang, T. and Zeng, Wangding and Liu, Wen and Liang, Wenfeng and Gao, Wenjun and Yu, Wenqin and Zhang, Wentao and Xiao, W. L. and An, Wei and Liu, Xiaodong and Wang, Xiaohan and Chen, Xiaokang and Nie, Xiaotao and Cheng, Xin and Liu, Xin and Xie, Xin and Liu, Xingchao and Yang, Xinyu and Li, Xinyuan and Su, Xuecheng and Lin, Xuheng and Li, X. Q. and Jin, Xiangyue and Shen, Xiaojin and Chen, Xiaosha and Sun, Xiaowen and Wang, Xiaoxiang and Song, Xinnan and Zhou, Xinyi and Wang, Xianzu and Shan, Xinxia and Li, Y. K. and Wang, Y. Q. and Wei, Y. X. and Zhang, Yang and Xu, Yanhong and Li, Yao and Zhao, Yao and Sun, Yaofeng and Wang, Yaohui and Yu, Yi and Zhang, Yichao and Shi, Yifan and Xiong, Yiliang and He, Ying and Piao, Yishi and Wang, Yisong and Tan, Yixuan and Ma, Yiyang and Liu, Yiyuan and Guo, Yongqiang and Ou, Yuan and Wang, Yuduan and Gong, Yue and Zou, Yuheng and He, Yujia and Xiong, Yunfan and Luo, Yuxiang and You, Yuxiang and Liu, Yuxuan and Zhou, Yuyang and Zhu, Y. X. and Huang, Yanping and Li, Yaohui and Zheng, Yi and Zhu, Yuchen and Ma, Yunxian and Tang, Ying and Zha, Yukun and Yan, Yuting and Ren, Z. Z. and Ren, Zehui and Sha, Zhangli and Fu, Zhe and Xu, Zhean and Xie, Zhenda and Zhang, Zhengyan and Hao, Zhewen and Ma, Zhicheng and Yan, Zhigang and Wu, Zhiyu and Gu, Zihui and Zhu, Zijia and Liu, Zijun and Li, Zilin and Xie, Ziwei and Song, Ziyang and Pan, Zizheng and Huang, Zhen and Xu, Zhipeng and Zhang, Zhongyu and Zhang, Zhen},
  year={2025},
}

@book{chatman1978story,
  title={Story and Discourse: Narrative Structure in Fiction and Film},
  author={Chatman, Seymour},
  year={1978},
  publisher={Cornell University Press}
}

@book{bal1997narratology,
  title={Narratology: Introduction to the Theory of Narrative},
  author={Bal, Mieke},
  year={1997},
  publisher={University of Toronto Press}
}

@article{cohn2013visual,
  title={Visual Narrative Structure},
  author={Cohn, Neil},
  journal={Cognitive Science},
  volume={37},
  number={3},
  pages={413--452},
  year={2013}
}

@incollection{ryan2005narrative,
  title={Narrative},
  author={Ryan, Marie-Laure},
  booktitle={Routledge Encyclopedia of Narrative Theory},
  editor={Herman, David and Jahn, Manfred and Ryan, Marie-Laure},
  year={2005},
  publisher={Routledge}
}

@article{hearst1997texttiling,
    title = "Text Tiling: Segmenting Text into Multi-paragraph Subtopic Passages",
    author = "Hearst, Marti A.",
    editor = "Hirschberg, Julia",
    journal = "Computational Linguistics",
    volume = "23",
    number = "1",
    year = "1997",
    address = "Cambridge, MA",
    publisher = "MIT Press",
    url = "https://aclanthology.org/J97-1003/",
    pages = "33--64"
}
